\documentclass{article} %
\usepackage{iclr2027_conference,times}
\usepackage{graphicx}

\usepackage{amsmath,amsfonts,bm}

\def\eqref#1{equation~\ref{#1}}

\def\1{\bm{1}}

\DeclareMathAlphabet{\mathsfit}{\encodingdefault}{\sfdefault}{m}{sl}
\SetMathAlphabet{\mathsfit}{bold}{\encodingdefault}{\sfdefault}{bx}{n}

\usepackage{hyperref}
\usepackage{url}
\usepackage{booktabs}
\usepackage{tabularx}
\usepackage{pifont}
\usepackage{enumitem}
\usepackage{caption}
\usepackage{wrapfig}
\usepackage[table]{xcolor}
\usepackage{fontawesome5}
\usepackage{hyperref}

\title{GenOlympiad: Benchmarking Frontier Image Models at Their Capability Boundaries}

\author{
Xi Hu \\
Australian National University \\
{\small\texttt{xi.hu@anu.edu.au}} \\
\And
Martin Bell \\
Canva \\
{\small\texttt{martinbell@canva.com}} \\
\And
Liang Zheng\thanks{Corresponding author.} \\
Australian National University \\
{\small\texttt{liang.zheng@anu.edu.au}}
}

\iclrfinalcopy
\begin{document}

\maketitle

\begin{abstract}
Text-to-image models have made tremendous progress and can produce  high-quality images. However, a question which would be asked in any other discipline is where the frontier models fall short. We introduce GenOlympiad, a benchmark which provides the most challenging text prompts from nine capability axes for stress-testing frontier text-to-image models. The assessed capabilities include generating large quantities, performing deep reasoning, ensuring hard physical consistency, following complex rules and so on. The prompts are designed such that prompt alignment is visually verifiable (\emph{e.g., counting}) instead of relying on subjectiveness (\emph{e.g., style}). Based on this, we develop an automated evaluation framework which decomposes the prompt into easy and hard questions, and for most hard questions, draws help lines to \textit{ground} evidence for accurate grading. 
We evaluate 20+ strong text-to-image models, including Nano Banana Pro, GPT-Image-2.5, Seedream-5.0 and so on. We report that even the best-performing T2I model (Nano Banana Pro) only achieves a strict score of 23.9 out of 100 and that coding models such as GPT-6 Astra have higher prompt alignment scores (but not as visually pleasing). We also derive insights in model evolution, alignment with Elo scores, and effectiveness of our grounding-based evaluation protocol. 
\par\smallskip
\noindent
\begin{tabular}{@{}lll@{}}
\faGithub & \textbf{Code} &
\href{https://github.com/GenOlympiad/GenOlympiad}
     {https://github.com/GenOlympiad/GenOlympiad} \\
\faGlobe & \textbf{Project Page} &
\href{https://genolympiad.github.io/}
     {https://genolympiad.github.io/}
\end{tabular}
\end{abstract}

\section{Introduction}
In artificial intelligence, benchmarks are introduced to expose frontier models to their capability boundaries, measure them, and seek solutions from the community. %
Some well known examples include ObjectNet~\citep{objectnet} for vision model generalization, Humanity's Last Exam~\citep{hle} for LLM question answering across subjects, SWE-bench~\citep{swebench} for automatic issue resolution, ARC-AGI benchmarks~\citep{arc1,arc2} for general intelligence, and FrontierMath~\citep{frontiermath} for testing AI on advanced mathematics. 

Frontier text-to-image models can now produce high quality images~\citep{imagenworld,qwenbench}. This is evidenced by both real user scoring, where platforms like Artificial Analysis Arena rank models by asking users to compare image pairs head-to-head, and strong scores on automatic benchmarks. GenEval~\citep{geneval} has risen from 0.61 when first introduced in 2023 to $\sim$0.90 for recent models. PRISM-Bench\citep{prismbench} score which measures image-text alignment and aesthetics is now over 85 by Nano Banana. 
However, these evaluation platforms either do not diagnose limitations of models (e.g., AA Arena) or see scores being saturated (e.g., GenEval) and thus no longer sit at capability boundaries. Therefore, we ask the question: \emph{when generated images already look high-quality, where do frontier image models still fail?}

We introduce GenOlympiad, an exam of 801 hardest prompts for frontier model evaluation, co-designed by AI and human. The prompts span nine capability axes, including quantity and precision, latent computation, physical consistency, spatial and 3D reasoning. To be hard, instead of adding more objects or textual conditions, the GenOlympiad prompts focus on hard  constraints that must be strictly satisfied. An example prompt is `a compass on a boat, with its needle pointing toward the setting sun': although frontier models can generate realistic lighting and textures, they usually fail to satisfy the constraint of `needle pointing towards the setting sun', which is what we assess. 

\begin{figure*}[t]
    \centering
    \includegraphics[
        width=\textwidth,
    ]{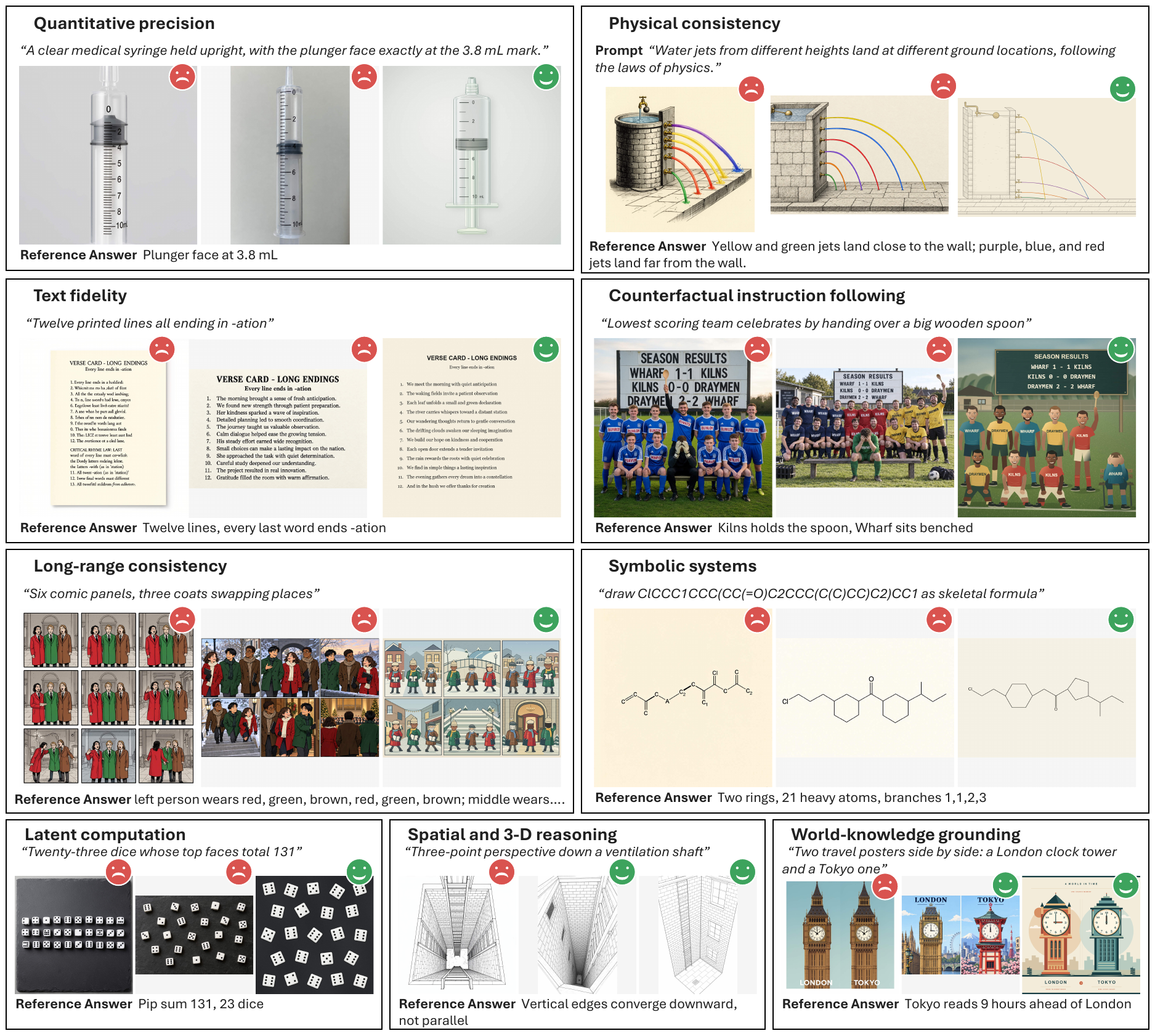}
    \caption{Image generation examples of the nine capability axes in GenOlympiad. In each capability, images from left to right are generated by FLUX.2 dev, GPT Image 2.5, and GPT-6 Astra, respectively. The prompts are shortened for demonstration. More examples are in Appendix Fig.~\ref{fig:wall}.}
    \label{fig:9axes}
\end{figure*}

Given each GenOlympiad prompt and an image generated by a frontier model, we design an automatic pipeline to score the generations, which focuses on constraint satisfaction instead of general alignment. Specifically, we do not ask the vision language model (VLM) `are there eight blue caps', because the answer can often be wrong. Instead,  we break it down to atomic tasks, \emph{e.g.}, detecting all the blue caps, drawing the bounding boxes on the image, and asking whether each bounding box contains a blue cap. Evaluation becomes more accurate and interpretable if we let the VLM perform such simple evidence extraction tasks and ground the task results (drawing bounding boxes and help lines). Our scoring function is deterministic, matching the extracted evidence (\emph{e.g.}, bounding boxes height, width, and coordinates) against the prompt. In addition, when grounding fails, we withhold the task instead of assigning an unreliable score.

We evaluate a suite of frontier models on GenOlympiad with our evaluation protocols. We report \emph{strict score} which requires all constraints in a task to be correct, and \emph{relaxed score} which gives a partial score for some constraints. Detailed definition is in Section~\ref{sec:deterministic_scoring}. Nano Banana Pro gains 23.9 in strict score and 40.0 in relaxed score out of 100, ranking first among T2I models, while the latest GPT-Image-2.5-sunburst ranks second at 14.9 in strict score and 29.6 in relaxed score. 
We also find that frontier closed models, e.g., the GPT-Image family, exhibit better GenOlympiad scores with their new releases, while open-weight image models have mixed results, suggesting that the latter may not have been trained at the boundary. In sum, this paper has the following main points. 

\begin{itemize}[leftmargin=*]
\item We introduce GenOlympiad, a test set of 801 hardest prompts to evaluate frontier models. 
\item We develop an automatic pipeline that provides robust evaluation for each prompt. 
\item Benchmarking a wide range of frontier models reveals that frontier models have poor performance and gives us rich insights in the boundary limitations and model development trend.
\end{itemize}

\section{Related work}

\paragraph{Benchmarks for text-to-image generation.}
Some popular benchmarks focus on object presence and attributes like color, relations or counting of a few objects~\citep{geneval, geneval2, countbench,echo4o, compbench++}. Others focus on creative images that rarely appear in the training set~\citep{conceptmix} or long prompt following, such as DPG-Bench~\citep{ella} and TIIF-Bench~\citep{tiifbench}. %
Recently, more challenging benchmarks are introduced by adding reasoning, \emph{e.g.}, Einstein's favorite instrument~\citep{genaibench,r2ibench,t2ireasonbench,wise,dscore,oneigbench}. Tasks in GenExam~\citep{genexam} reflect various disciplines such as  history, music, and biology, while MathGen~\citep{mathgen} focuses on mathematics content, \emph{e.g.}, angles, function, set and counting tasks with many objects. Prompts in Qwen-Image-Bench \citep{qwenbench} and ImagenWorld \citep{imagenworld} instead reflect real design demand in games, products, fashion, \textit{etc}. Earlier benchmarks like GenEval  have already been saturated to 90\%. 
Frontier models also made rapid improvement on recent benchmark, \emph{e.g.}, GenExam reports a 74.6\% strict score for GPT-Image-2; while MathGen report 53.7\% pass rate (similar to strict score) for Nano Banana Pro. In comparison, GenOlympiad curates the hardest prompts and only sees 23.9\% and 14.9\% of Nano Banana Pro and GPT-Image-2.5 Sunburst.

\paragraph{Evaluation of generated images.}
Early metrics like Inception Score (IS) \citep{is} and Fréchet Inception Distance (FID) \citep{fid} measure how realistic images are. Later metrics focus more on image-text alignment. CLIPScore~\citep{clipscore} calculates the similarity between prompt and image, which is not strongly correlated with human judgement~\citep{geneval}. Human based metrics like Elo in Artificial Analysis Image Arena was later adopted, but such metrics are too expensive to use and do not diagnose limitations of models. VQA-based automatic metrics mitigate these problems \citep{tifa,dsg,vqascore} by directly applying VQA models to image-question pairs. Other evaluation methods use vision tools to obtain important evidence about texts or objects, \textit{e.g.},  
GenEval~\citep{geneval} uses detection models; %
MathGen~\citep{mathgen} uses OCR,  contour detection, \textit{etc} to extract text, angles, \textit{etc}. GenOlympiad is similar to GenExam/TIFA/DA-Score in breaking the prompt into questions, but is different in using visual grounding in evaluation. GenOlympiad also shares similarity with GenEval/MathGen and thinking with images~\citep{thinkingwithimage} in using vision tool for evidence collection; our key difference is using an iterative process to improve grounding (Table \ref{tab:human_agreement}). %

\section{The GenOlympiad benchmark}
\label{sec:benchmark}

\subsection{What composes a task?}
\label{sec:whatcomposesatask}

Each GenOlympiad task contains a natural-language prompt and a bespoke evaluation program, both of which are provided during evaluation. A \textbf{prompt} describes the image to be generated. Each prompt has at least one aspect which is very hard to generate, \emph{e.g.}, 35 objects. The brainstorming of the 801 prompts was performed by the authors with the help of AI, details is in Appendix~\ref{sec:howwebuild}.

The \textbf{evaluation program} includes 
easy questions and hard questions. It also includes the pipelines for evidence grounding and scoring (Section \ref{sec:framework}).  %
The program is created by Claude Opus 5 with human instructions (for details see Appendix~\ref{sec:howwebuild}). As shown in Fig.~\ref{fig:hardandeasy}, easy questions look like 
`are there many dice in this photo?' and `are the top faces of the dice clearly visible?'
Example hard questions can be `Do these dice sum up to 131?' and `How many dice in this picture?' %

\begin{figure}[t]
    \centering
    \includegraphics[width=\textwidth]{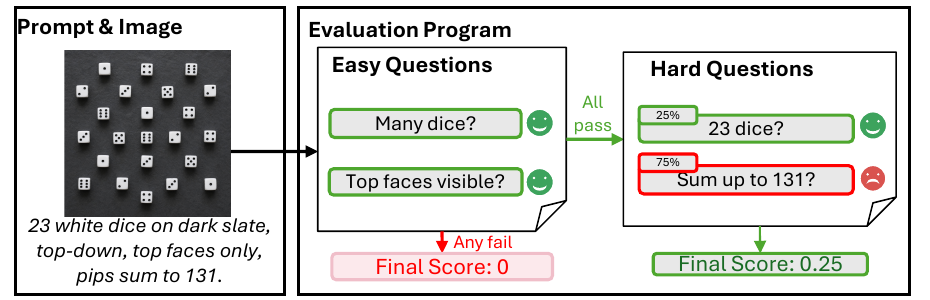}
    \caption{
        Example of a GenOlympiad prompt and its evaluation program. \textbf{Left}: prompt and a generated image. \textbf{Right}: Easy and hard questions compose  evaluation program. Easy questions are evaluated first as a `gating' check: if any fails, the final score will be zero. If all pass, the program will mark hard questions, where each hard question has partial mark with different weight.
    }
    \label{fig:hardandeasy}
\end{figure}

\begin{figure}[t]
    \centering
    \includegraphics[width=\textwidth]{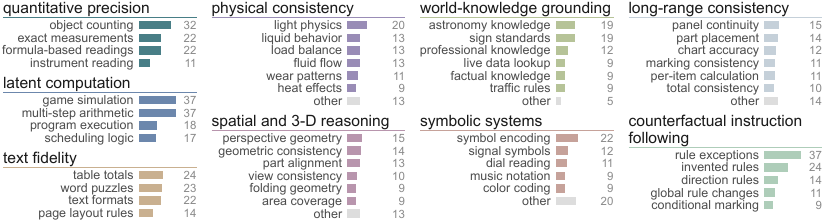}
    \caption{
        Topic distribution of GenOlympiad on each axis. We show the number of tasks in each topic. A detailed distribution can be seen in Appendix Fig.~\ref{fig:axistopic}
    }
    \label{fig:topic}
\end{figure}
\subsection{What does GenOlympiad consist of?}

The 801 tasks in GenOlympiad overall evaluate nine capability axes below (topic distribution in Fig.~\ref{fig:topic}). More example prompts and success/failure images are shown in Appendix Fig.~\ref{fig:9axes} and Fig.~\ref{fig:wall}. 

\begin{itemize}[leftmargin=*]
\item {Quantity and precision}: to generate a large number of objects, exact ratio between objects, precise angles, and precise readings of instrument like ruler.
\item {Latent computation}: computation outcome does not appear in the prompt, requiring the system to analyze the state for generation, \emph{e.g.}, \textit{23 dice which sum up to 131}.
\item {Physical consistency}: to follow real-world physical constraints like gravity, optics, and stress, \emph{e.g.}, \textit{water jets from different heights land at different ground locations}.
\item {Spatial and 3D reasoning}: examine the correctness of perspective relations and the expand diagram of 3D objects, \emph{e.g.}, \textit{a picture should be drawn in 3-point perspective}.
\item {Text fidelity}: readable text with constraints like font size and language habits like inverted question mark in Spanish, \emph{e.g.}, \textit{a paper printed with 12 lines of rhyme, each line ending with -ation}.
\item {Symbolic systems}: follow symbols in circuit, molecular graph, music sheet, \textit{etc}, \emph{e.g.}, \textit{draw ClCCC1CCC(CC(=O)C2CCC(C(C)CC)C2)CC1 as a line-angle (skeletal) structural formula}.
\item {Long-range consistency}: parts of an image, even parts far apart, should be consistent, \emph{e.g.}, consistency of character in hats, clothes and rings in a multi-panel comic.
\item {Counterfactual instruction following}: prompt is opposite to commonsense knowledge, \emph{e.g.}, \textit{a football team that gets the lowest score celebrate for its loss}.
\item {World-knowledge grounding}: to follow commonsense knowledge \emph{e.g.}, flags of countries; real-time information \emph{e.g.}, current air quality; The time difference between two specific countries should be correct.
\end{itemize}

\subsection{Evaluation framework}
\label{sec:framework}

We design an automatic pipeline, which runs the \textit{evaluation program} defined in Section \ref{sec:whatcomposesatask}, to grade how well a generated image follows the prompt.

\textbf{Easy questions easily asked.} We use a TIFA-style method for easy question evaluation: the judge VLM takes the easy question and the generated image as input and answers yes or no. Asking once is enough, since the agreement with human answers is 94.7\%, and majority voting over 3 independent queries only slightly improves it to 95.3\% at extra cost. %
As shown in Fig.~\ref{fig:hardandeasy}, only after all easy questions have been passed will we proceed with the hard questions; otherwise, we directly mark 0.

\begin{figure}[t]
    \centering
    \includegraphics[width=\textwidth]{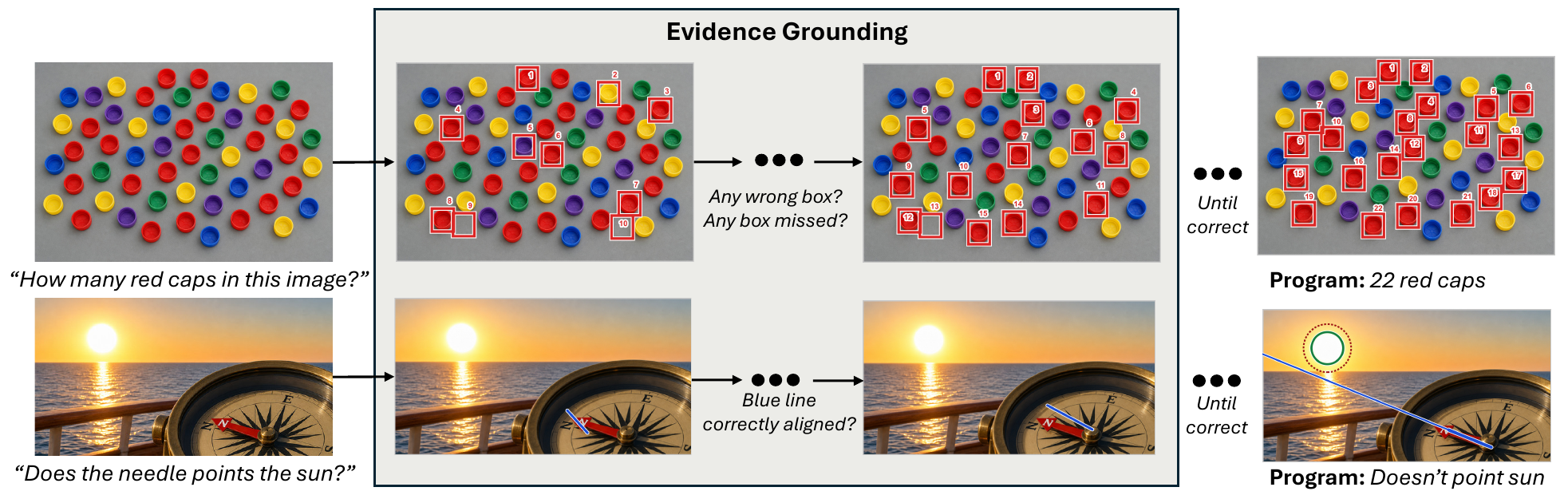}
    \caption{
    Pipeline of evidence grounding. Given a hard question and the generated image (left), we draw helping lines on the image, \emph{e.g.}, bounding boxes for counting problems, boundaries for the sun, and extension lines for the compass pointer. The type of helping line is decided in the evaluation program. To make the lines accurate, we design an iterative process (middle), where we use Gemini 2.5 Flash to decide whether each line serves the purpose, until success or running out of budget. The grounding result (right) is finally used to decide whether the hard constraint is satisfied. 
    }
    \label{fig:eval}
\end{figure}

\textbf{Grounding visual evidence for hard questions.} 557 out of 801 tasks in GenOlympiad require evidence grounding. 
Depending on questions, we draw different helping lines on the generated image. As shown in Fig. \ref{fig:eval}, for counting questions, we aim to draw bounding boxes of objects of interest; for the compass-sun question, we aim to draw an extension line of the compass pointer and contour of the sun. There is no guarantee that grounding is successful at the first try, so after each try, we ask the VLM whether grounding is successful, such as `does the blue line correctly overlay with the compass pointer?' If  successful, then grounding is complete; if not successful, we repeat the grounding step until reaching maximum trial times. The grounded evidence, \emph{e.g.}, bounding boxes and lines, is then processed by a deterministic program. The program performs bespoke operations, such as counting bounding boxes, calculating the sum of dice or checking whether the compass-pointer line intersects the contour of the sun, to determine whether the hard question is satisfied.

\begin{figure}[t]
    \centering
    \includegraphics[width=\textwidth]{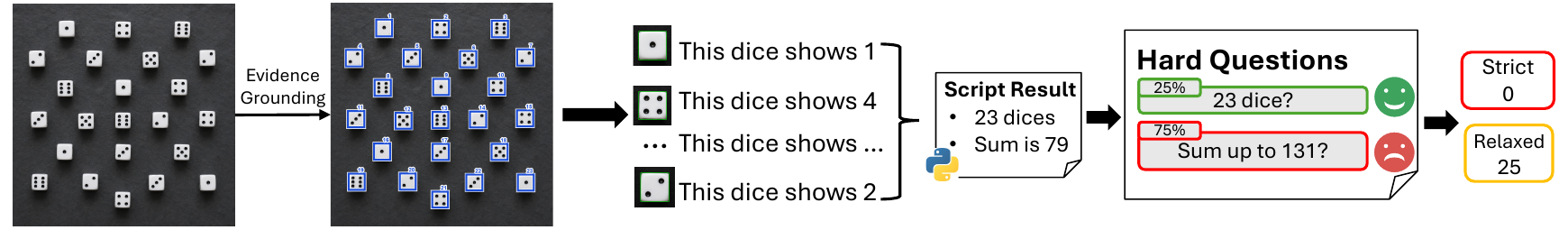}
    \caption{
        An example pipeline of deterministic scoring. We write the grounding result into a JSON state file, which, for each die, lists the bounding box height, width, and center coordinates, and the displayed number. We use a deterministic Python script to count the number of dice and the sum of the numbers for grading. The entire deterministic scoring pipeline is provided in the evaluation program. Prompt is the same as Fig. \ref{fig:hardandeasy}. 
    }
    \label{fig:deterministic}
\end{figure}

\label{sec:deterministic_scoring}
\textbf{Deterministic scoring.} 
After each hard question has been verified, the scoring method will count the number of hard questions that are successfully reflected in the generated image. 
For example, there are two hard questions for Fig.~\ref{fig:deterministic}: (1) is the number of dice 23; (2) do these dice sum up to 131?
Following GenExam~\citep{genexam}, we report two scores. In our `strict' scoring, only when all the hard questions are successfully passed will we give a mark of 100. In the `relaxed' setting, different hard questions have different weights, and the weight sum of the successful ones is the final score. The strict score is the percentage of the scored tasks that pass all the hard questions.

\textbf{Grounding failure.}  If grounding is unsuccessful after the maximum number of trials, we remove the task rather than force the grader to produce a potentially unreliable score. The failure scenarios may be: an object is too small; object color is very close to the background; many objects overlay each other and can not be clearly located. %
Grounding may also fail if a rendered text is not recognizable or the generated objects do not match the objects expected by the evaluation program. We rank models according to their grounding failure rate in Fig.~\ref{fig:witheld}.

\section{Experiments}
\label{sec:experiments}

We perform extensive evaluation and insight analysis of frontier image models on GenOlympiad. In particular, we investigate the following research questions. 
\begin{itemize}[leftmargin=*]
\item What do comprehensive benchmarking results look like (Section~\ref{sec:results}, Table~\ref{tab:main_results}, Fig.~\ref{fig:witheld})? Do frontier models have improved GenOlympiad scores over releases? (Section~\ref{sec:analysis}, Fig.~\ref{fig:serues_radars}, Appendix Fig.~\ref{fig:open_series})
\item What is the correlation of GenOlympiad scores with human ratings on Elo Arenas? (Section~\ref{sec:analysis}, Fig.~\ref{fig:arenacorrelation}, and Appendix Fig.~\ref{fig:witheld_corr},~\ref{fig:other_corr})
\item How strongly does GenOlympiad evaluation results agree with human ratings? (Section~\ref{sec:eval_validation}, Table~\ref{tab:human_agreement}, and Appendix Tables~\ref{tab:human_pass_fail},~\ref{tab:reader_consistency})
\end{itemize}

\subsection{Experimental setup}
\label{sec:setup}
We benchmark the following T2I models: GPT-Image 1, 1.5, 2, and 2.5 Sunburst~\citep{gptimage1,gptimage15,gptimage2,gptimage25}; Nano Banana, Nano Banana Pro, and Nano Banana 2~\citep{nanobanana,nanobananapro,nanobanana2}; Seedream 4.5 and 5.0 Pro~\citep{seedream45,seedream5}; FLUX.2 Pro, dev, and Klein~\citep{flux2}; FLUX.1 dev and Schnell~\citep{flux1}; Stable Diffusion 3.5~\citep{sd3,sd35}; SANA~\citep{sana}; and Qwen-Image-20B~\citep{qwenimage}.

We evaluate open-weight models ranging from 0.6B to 20B parameters on the same server with a single NVIDIA A100 80GB GPU. For each model family, we configure the context length of the primary text-conditioning encoder to make the full prompt visible whenever possible. If prompt length exceeds the context length, it will neither generate nor be calculated into the overall score. %

We also benchmark two coding models, \emph{i.e.}, GPT-6 Astra~\citep{gpt6astra}, GPT-5.6 Sol~\citep{gpt56} in medium effort with internet accessibility. We insert a system prompt (See Appendix~\ref{sec:prompt}) to let the coding models write a single HTML document. The content on this document is rendered programmatically by JavaScript. %
We evaluate image aesthetics using the same method and setup as PRISM-Bench~\citep{prismbench}, where the evaluator is GPT-4.1~\citep{gpt41}. %

Prompt engineering (PE), which preserves the original constraints while improving T2I performance, is evaluated on GPT-Image-2.5. We use PromptEnhancer-32B~\citep{PE32B} and GPT-5.6 Sol. For GPT-5.6 Sol, we use a fixed prompt from Tailored Visions~\citep{PEStrongModel}. %

\subsection{Benchmarking results}
\label{sec:results}
\begin{table}[t]
\centering
\caption{
GenOlympiad results for closed-source T2I models, coding models, prompt engineering and open-source T2I models across nine capability axes. Best result in each column and each category is highlighted in \textbf{bold}. C-Fact: counterfactual instruction following; Symbol: symbolic systems; World: world-knowledge grounding; Phys: physical consistency; Quant: quantity and precision; Spatial: spatial and 3D reasoning; Latent: latent computation; Long: long-range consistency. Overall is the mean score over all scored tasks of a model. Withheld tasks are excluded from both axis and overall scores (see Fig.~\ref{fig:witheld}). S and R are strict and relaxed scores.
}
\label{tab:main_results}

\scriptsize
\setlength{\tabcolsep}{0pt}
\renewcommand{\arraystretch}{1.0}
\begin{tabular*}{\textwidth}{
@{\extracolsep{\fill}}
l@{\hspace{1.5pt}}|
cc cc cc cc cc cc cc cc cc
@{\hspace{5pt}}|
cc
@{}
}
\toprule

\textbf{Model}
& \multicolumn{2}{c}{\textbf{Text}}
& \multicolumn{2}{c}{\textbf{C-Fact}}
& \multicolumn{2}{c}{\textbf{Symbol}}
& \multicolumn{2}{c}{\textbf{World}}
& \multicolumn{2}{c}{\textbf{Phys}}
& \multicolumn{2}{c}{\textbf{Quant}}
& \multicolumn{2}{c}{\textbf{Spatial}}
& \multicolumn{2}{c}{\textbf{Latent}}
& \multicolumn{2}{c}{\textbf{Long}}
& \multicolumn{2}{c}{\textbf{Overall}}
\\

&
\textbf{S} & \textbf{R}
& \textbf{S} & \textbf{R}
& \textbf{S} & \textbf{R}
& \textbf{S} & \textbf{R}
& \textbf{S} & \textbf{R}
& \textbf{S} & \textbf{R}
& \textbf{S} & \textbf{R}
& \textbf{S} & \textbf{R}
& \textbf{S} & \textbf{R}
& \textbf{S} & \textbf{R}
\\
\midrule

\rowcolor{gray!15}
\multicolumn{21}{c}{\textbf{\textit{Coding Models}}} \\
GPT-6 Astra
& \textbf{65.4} & \textbf{74.1}
& \textbf{38.4} & \textbf{50.2}
& \textbf{45.5} & \textbf{56.4}
& \textbf{27.5} & \textbf{46.9}
& 25.6 & \textbf{43.0}
& \textbf{40.8} & \textbf{54.0}
& 26.7 & 48.3
& \textbf{58.4} & \textbf{73.7}
& \textbf{30.0} & \textbf{53.1}
& \textbf{40.5} & \textbf{56.1}
\\
GPT-5.6 Sol 
& 62.5 & 72.4 
& 32.2 & 43.3 
& 42.3 & 55.5 
& 26.0 & 43.5 
& \textbf{27.7} & 41.6 
& 31.6 & 45.8 
& \textbf{28.2} & \textbf{49.4}
& 53.3 & 69.1 
& 28.4 & 46.7 
& 37.5 & 52.4 \\

\rowcolor{gray!15}
\multicolumn{21}{c}{\textbf{\textit{Prompt Engineering on GPT-Image-2.5}}} \\
GPT 5.6 Sol
& \textbf{62.2} & \textbf{71.7}
& \textbf{33.0} & \textbf{48.7}
& \textbf{30.4} & \textbf{44.7}
& \textbf{17.8} & \textbf{36.2}
& \textbf{25.0} & \textbf{39.4}
& \textbf{24.3} & \textbf{40.8}
& \textbf{12.5} & \textbf{34.3}
& \textbf{55.8} & \textbf{72.3}
& \textbf{27.2} & \textbf{43.2}
& \textbf{33.0} & \textbf{48.9}
\\
PromptEnhancer
& 26.9 & 38.7
& 21.3 & 33.9
& 19.8 & 30.7
& 6.8 & 19.0
& 8.3 & 24.5
& 16.7 & 27.7
& 8.9 & 26.8
& 15.2 & 30.6
& 11.0 & 22.3
& 15.1 & 28.4
\\

\rowcolor{gray!15}
\multicolumn{21}{c}{\textbf{\textit{Closed-source T2I Models}}} \\
Nano Banana Pro
& \textbf{41.5} & \textbf{55.8}
& \textbf{27.8} & \textbf{39.3}
& \textbf{25.6} & \textbf{38.6}
& 10.5 & \textbf{29.5}
& 13.1 & \textbf{33.8}
& \textbf{17.1} & \textbf{27.0}
& 11.2 & 30.7
& \textbf{38.1} & \textbf{56.0}
& \textbf{24.4} & \textbf{43.2}
& \textbf{23.9} & \textbf{40.0}
\\
GPT-Image-2.5
& 30.4 & 41.8
& 23.0 & 36.4
& 20.5 & 34.8
& 6.8 & 24.1
& 8.1 & 23.7
& 13.2 & 25.0
& \textbf{18.4} & \textbf{34.6}
& 6.5 & 21.9
& 9.8 & 25.9
& 14.9 & 29.6
\\
Seedream 5.0 Pro
& 19.0 & 30.8
& 24.1 & 36.0
& 18.9 & 32.7
& 2.5 & 16.6
& \textbf{13.8} & 30.8
& 10.4 & 20.7
& 9.9 & 23.7
& 20.8 & 34.7
& 13.7 & 26.6
& 15.1 & 28.4
\\
Nano Banana 2
& 19.5 & 28.5
& 13.4 & 27.3
& 6.6 & 18.4
& \textbf{13.9} & 27.0
& 8.9 & 23.0
& 6.8 & 18.3
& 8.1 & 20.9
& 16.2 & 32.5
& 15.9 & 29.3
& 12.4 & 25.4
\\
GPT-Image-2
& 20.8 & 32.8
& 14.0 & 26.7
& 17.5 & 29.6
& 5.4 & 17.2
& 6.1 & 20.3
& 5.2 & 17.3
& 10.1 & 24.9
& 5.6 & 18.2
& 8.8 & 25.1
& 10.2 & 23.4
\\
GPT-Image-1.5
& 10.1 & 19.8
& 10.1 & 19.9
& 6.2 & 12.9
& 2.9 & 15.1
& 3.2 & 15.7
& 5.6 & 15.0
& 10.7 & 19.3
& 2.9 & 12.5
& 3.5 & 12.7
& 6.0 & 15.8
\\
Nano Banana
& 7.8 & 13.5
& 5.9 & 16.0
& 7.6 & 18.5
& 4.1 & 14.3
& 6.3 & 21.6
& 1.3 & 6.0
& 5.5 & 12.0
& 8.7 & 18.6
& 1.3 & 15.0
& 5.5 & 15.3
\\
FLUX.2 pro
& 7.1 & 12.0
& 6.0 & 15.9
& 5.0 & 12.9
& 1.6 & 9.5
& 6.9 & 17.4
& 2.7 & 12.9
& 1.8 & 9.9
& 17.3 & 24.8
& 1.9 & 14.2
& 6.3 & 15.0
\\
GPT-Image-1
& 11.4 & 20.1
& 7.0 & 15.2
& 10.5 & 21.1
& 3.9 & 15.1
& 5.1 & 15.4
& 2.7 & 11.4
& 1.3 & 12.2
& 2.9 & 12.6
& 2.6 & 11.8
& 5.2 & 14.9
\\
Seedream-4.5
& 3.0 & 8.3
& 0.0 & 10.2
& 0.0 & 5.7
& 0.0 & 8.8
& 8.5 & 16.6
& 5.7 & 12.3
& 0.0 & 10.4
& 5.8 & 12.5
& 1.8 & 9.2
& 3.0 & 10.6
\\

\rowcolor{gray!15}
\multicolumn{21}{c}{\textbf{\textit{Open-source T2I Models (1024$\times$1024, default settings, see Table~\ref{tab:open_source_settings})}}} \\
FLUX.2 dev
& 1.4 & 4.1
& 3.6 & \textbf{9.3}
& \textbf{4.1} & 9.3
& 3.0 & 11.4
& 10.0 & 18.4
& 1.4 & 7.5
& 5.6 & \textbf{12.8}
& 1.1 & \textbf{7.8}
& 1.3 & \textbf{12.1}
& \textbf{3.5} & \textbf{10.3}
\\
FLUX.2 klein 9B
& 0.0 & \textbf{4.2}
& 2.5 & 9.2
& 2.8 & 8.8
& 3.1 & 11.0
& \textbf{10.1} & \textbf{19.0}
& 2.7 & 8.4
& 2.7 & 10.5
& 2.0 & 6.6
& 0.0 & 9.7
& 2.9 & 9.6
\\
FLUX.2 klein 4B
& 0.0 & 1.8
& \textbf{3.9} & 8.3
& 2.7 & \textbf{10.2}
& \textbf{6.6} & \textbf{17.5}
& 5.3 & 11.7
& 0.0 & 7.7
& \textbf{7.1} & 12.4
& 0.0 & 6.1
& 2.7 & 7.1
& 2.9 & 8.9
\\
FLUX.1 dev
& \textbf{1.7} & 2.2
& 2.9 & 6.7
& 3.2 & 6.5
& 5.8 & 11.3
& 7.4 & 15.1
& \textbf{3.3} & \textbf{9.7}
& 3.3 & 6.1
& \textbf{3.6} & 6.6
& 0.0 & 4.8
& \textbf{3.5} & 7.7
\\
FLUX.1 schnell
& 1.6 & 3.9
& 2.9 & 7.2
& 0.0 & 4.0
& 3.1 & 9.0
& 8.3 & 14.9
& 3.2 & 7.0
& 2.9 & 12.4
& 0.0 & 2.9
& \textbf{3.4} & 7.1
& 2.8 & 7.6
\\ 

Qwen-Image 20B 
& 0.0 & 1.5 
& 0.0 & 2.2 
& 1.5 & 4.5 
& 6.2 & 11.0 
& 7.8 & 13.1 
& 3.1 & 7.1 
& 1.6 & 7.4 
& 0.0 & 2.7 
& 1.6 & 5.3 
& 2.2 & 5.7 \\
SD3.5 large 
& 0.0 & 0.6 
& 1.5 & 3.8 
& 0.0 & 2.8 
& 5.4 & 8.9 
& 1.5 & 5.3 
& 1.6 & 3.1 
& 1.5 & 9.7 
& 0.0 & 2.9 
& 1.4 & 4.7 
& 1.3 & 4.5 \\
SD3.5 medium 
& 0.0 & 0.8 
& 0.0 & 1.7 
& 1.6 & 3.1 
& 3.8 & 8.8 
& 2.8 & 7.8 
& 1.6 & 5.0 
& 0.0 & 4.4 
& 0.0 & 3.6 
& 0.0 & 5.8 
& 1.0 & 4.5 \\
Sana 4.8B 
& 0.0 & 0.7 
& 0.0 & 1.9 
& 0.0 & 2.7 
& 1.9 & 5.2 
& 6.0 & 10.5 
& 0.0 & 3.6 
& 1.7 & 5.3 
& 0.0 & 4.4 
& 0.0 & 2.4 
& 1.0 & 4.1 \\
Sana 0.6B 
& 0.0 & 0.6 
& 0.0 & 1.6 
& 1.7 & 3.0 
& 1.9 & 5.5 
& 4.3 & 7.0 
& 1.8 & 5.7 
& 1.6 & 6.7 
& 0.0 & 3.4 
& 0.0 & 2.4 
& 1.2 & 3.9 \\
Sana 1.6B 
& 1.5 & 1.7 
& 3.4 & 4.0 
& 1.7 & 2.6 
& 1.7 & 6.7 
& 3.1 & 5.8 
& 0.0 & 1.1 
& 1.8 & 6.3 
& 0.0 & 2.7 
& 0.0 & 2.0 
& 1.4 & 3.6 \\
\bottomrule
\end{tabular*}
\end{table}

\textbf{Comparing closed-source T2I models.} 
As shown in Table \ref{tab:main_results}, Nano Banana Pro obtained 40.0 relaxed score and 23.9 strict score and performs the best in text fidelity, counterfactual instruction following, symbolic systems and so on. GPT-Image-2.5 is on par with Seedream 5.0 Pro and Nano Banana 2. In addition, Seedream 5.0 Pro in latent computation and physical consistency performs better than all the GPT-Image series. %

\textbf{Prompt engineering (PE) analysis.} We test whether prompt rewriting is useful for GenOlympiad. We use GPT-Image-2.5 as the base model. We compare PromptEnhancer-32B and GPT-5.6 Sol for prompt rewriting while keeping the image model and evaluator unchanged. Results are shown in Table \ref{tab:main_results}. PromptEnhancer-32B brings nearly no overall improvement over GPT-Image-2.5, while GPT-5.6 Sol increases the strict score and relaxed score from 14.9 to 33.0 and from 29.6 to 48.9, respectively. The gain is significant in latent computation and text fidelity and very limited in spatial and 3D reasoning. This demonstrates the usefulness of PE in T2I system design. %
\begin{figure}[t]
    \centering
    \includegraphics[width=\textwidth]{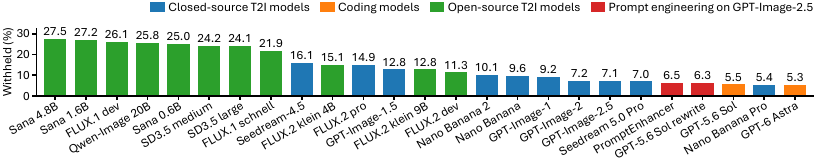}
    \caption{
    Grounding failure rate for different models. Lower is better. Closed-source T2I models are shown in blue, coding models in orange, and open-weight T2I models in green. Prompt Engineering in red. The Spearman correlation coefficient between failure rate and Olympiad strict scores is -0.92. It indicates that stronger models generally see lower grounding failures in our evaluation pipeline. %
    }
    \label{fig:witheld}
\end{figure}

\begin{figure}[t]
    \centering
    \includegraphics[width=\textwidth]{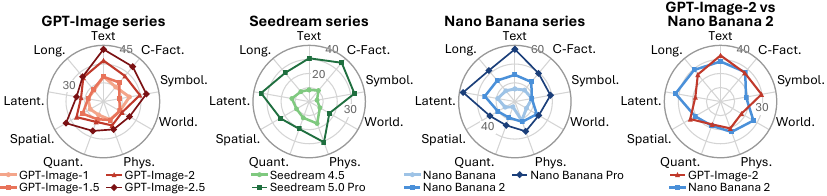}
    \caption{
    Performance of GPT, Seedream and Nano Banana series across nine axes. And the performance between GPT-Image-2 and Nano Banana 2. The performance improves as stronger models are released. This might not be the case for less strong models (see Appendix Fig.~\ref{fig:open_series})
    }
    \label{fig:serues_radars}
\end{figure}

\begin{table*}[t]
\centering

\begin{minipage}[t]{0.44\textwidth}
\centering
\small
\caption{
Agreement between our evaluation and human judgment. The first three rows show human agreement on different types of questions. %
Last two rows compare evaluation with and without evidence grounding on 117 hard questions requiring grounding.
}
\label{tab:human_agreement}
\begin{tabular}{@{}lr@{}}
\toprule
\textbf{Setting} & \textbf{Agreement} \\
\midrule
All questions & 95.3\% \\
Easy questions & 94.7\% \\
Hard questions & 95.7\% \\
\midrule
Hard w/o grounding & 68.4\% \\
Hard w/ grounding & 91.5\% \\
\bottomrule
\end{tabular}
\end{minipage}
\hfill
\begin{minipage}[t]{0.52\textwidth}
\centering
\small
\caption{
GenOlympiad scores versus aesthetic quality across models. Aesthetic scores are rated from 0 to 10 by GPT-4.1 following PRISM-Bench. Higher is better. Coding models are marked by $^\dagger$.
}
\label{tab:coding_aesthetics}
\setlength{\tabcolsep}{6pt}
\begin{tabular}{lccc}
\toprule
& \multicolumn{2}{c}{\textbf{GenOlympiad}} & \textbf{Aesthetic} \\
\cmidrule(lr){2-3}
\textbf{Model} & \textbf{Strict} & \textbf{Relaxed} & \textbf{Score} \\
\midrule
GPT-Image-2.5     & 14.9 & 29.6 & \textbf{5.51} \\
Nano Banana Pro  & 23.9 & 40.0 & 5.08 \\
Seedream 5.0 Pro & 15.1 & 28.4 & 4.62 \\
GPT-6 Astra$^\dagger$    & \textbf{40.5} & \textbf{56.1} & 4.61 \\
GPT-5.6 Sol$^\dagger$    & 37.5 & 52.4 & 4.61 \\
Nano Banana      & 5.5  & 15.3 & 2.99 \\
\bottomrule
\end{tabular}
\end{minipage}

\end{table*}

\textbf{Coding models and aesthetics.}
Coding models generally have higher GenOlympiad scores than T2I models. For example, the strict scores of GPT-6 Astra and Nano Banana Pro are 40.5 and 23.9, respectively. This is because our evaluation focuses on reasoning, where these frontier models are strong. On the other hand, it is also expected that the aesthetic score of coding models is often lower than the strong T2I models (refer Table \ref{tab:coding_aesthetics} and Appendix Fig.~\ref{fig:aesthetic}). For example, the aesthetic scores of GPT-5.6 Sol and GPT-Image-2.5 are 4.61 and 5.51, respectively. 

\textbf{Case study: GPT-Image-2 \textit{vs}. Nano Banana 2.} They have comparable performance as shown in Table \ref{tab:main_results}. %
GPT-Image-2 scores higher on text fidelity, symbolic systems, and 3D reasoning, while Nano Banana 2 is better on world-knowledge grounding, physical consistency, latent computation, and long-range consistency. 
It suggests that they were trained on different data. Meanwhile, a lower overall score does not necessarily mean that a model has lower scores on every capability axis. %

\subsection{Further Analysis and Insights}
\label{sec:analysis}
\textbf{Comparing grounding failure rates across models.} As discussed in Section \ref{sec:framework}, grounding failures happen when some generated content cannot be grounded by Gemini 2.5 Flash. This does not always mean bad generation: occluded objects may cause failure but may not be bad. In Fig. \ref{fig:witheld}, we compare the 25 models of their grounding failures and find that their ranking highly correlates with their GenOlympiad scores (strict) with a Spearman coefficient of -0.92. It suggests that a stronger model also tends to produce more easily grounded objects, which could be a result of post-training. 

\textbf{Do frontier models improve on GenOlympiad across releases?} We evaluate different releases of frontier models \emph{e.g.}, GPT-Image series, Seedream series, Nano Banana series. Fig.~\ref{fig:serues_radars} visualizes Table~\ref{tab:main_results} as radar charts.. Later releases generally achieve higher scores, but the improvements are highly uneven across axes. Some capabilities, such as text, spatial reasoning, and latent computation, improve substantially, while others, such as world knowledge, show much slower progress. %

\begin{figure}[t]
    \centering
    \includegraphics[
        width=\textwidth,
    ]{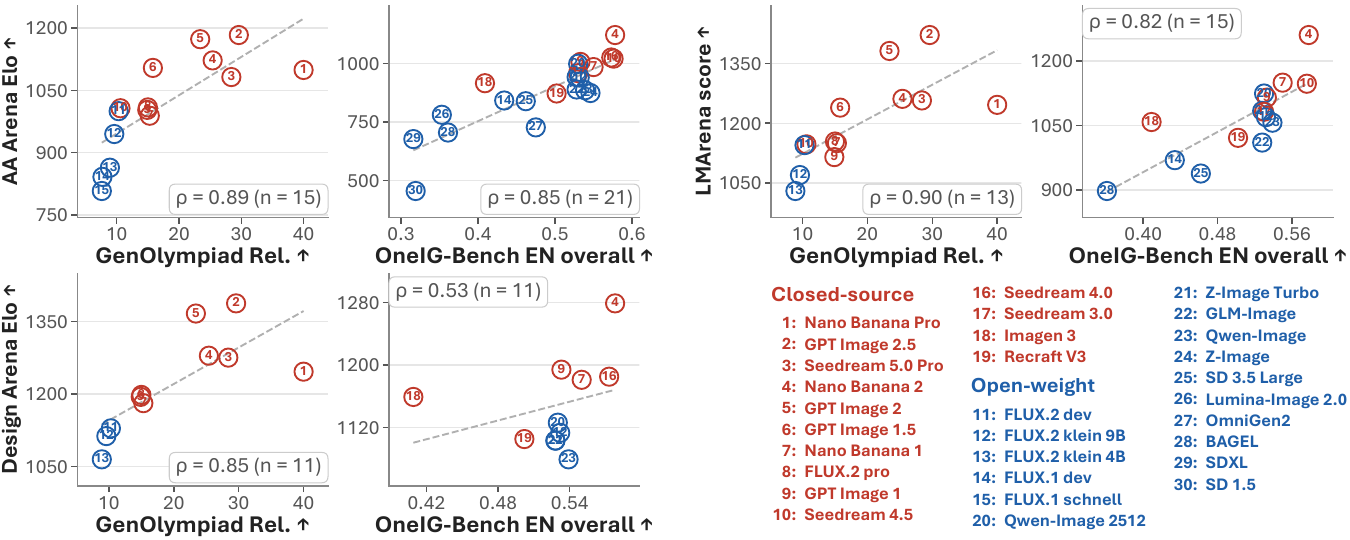}
    \caption{Correlation between GenOlympiad relaxed scores and three human-preference-based arena scores. OneIG-Bench is included for comparison. GenOlympiad seems to be a good proxy of Elo scores. Results for other benchmarks are shown in Appendix Fig.~\ref{fig:other_corr}.}
    \label{fig:arenacorrelation}
\end{figure}

\textbf{Correlation of GenOlympiad scores with human-based arena scores.}
We also analyze the correlation between GenOlympiad scores with human-based arena scores. The latter includes Image Arena, LMArena and Design Arena. Correlation results are shown in Fig. \ref{fig:arenacorrelation}. Although not intended, we observe strong correlation between GenOlympiad score and the arena scores. In other words, models that human prefers often obtain higher scores on GenOlympiad. On the other hand, outliers do exist. For example, Nano Banana Pro has much higher GenOlympiad score but only ranks in the middle of the three arenas. It indicates that differences exist between GenOlympiad and arenas. %

\subsection{Evaluator validation}
\label{sec:eval_validation}

\textbf{Agreement with human judgment.} 
 We check how the automatic evaluation protocol agrees with humans. %
 So we recruited 41 annotators on Prolific and collected human answers for 365 questions consisting of 131 easy questions and 234 hard questions. The evaluated images were generated by five T2I models and randomly selected. Annotators see the same generated image and the same questions in our evaluation program but do not see the model name, the VLM answers, or the final score. Each question was assigned to 3 annotators, and we use the majority answer as the human answer. The human answers are then sent to the same {deterministic scoring program} described in Section~\ref{sec:deterministic_scoring} for the final decision, \emph{e.g.}, the true/false answer of `Do these dice sum up to 131?'.  %
 
From Table \ref{tab:human_agreement}, we have two observations. First, the pass/fail agreement between human and automatic evaluation is 95.3\%(348/365). The agreement is 94.7\% (124/131) on easy questions and 95.7\% (224/234) on hard. Second, evidence grounding is effective. We test this on 117 questions that require evidence grounding, and here the agreement means that the answer is the same as the human answer. If we directly ask VLM without grounding, the agreement between human and VLM is 68.4\% (80/117). When we perform evidence grounding, the agreement becomes 91.5\% (107/117).  %

\begin{wraptable}{r}{0.43\textwidth}
\vspace{-13pt}
\centering
\small
\caption{Pairwise agreement between three different VLM judges on the same 365 set of human evaluation questions.}
\label{tab:reader_consistency}
\begin{tabular}{@{}lc@{}}
\toprule
\textbf{VLM pair} & \textbf{Agreement} \\
\midrule
Gemini 2.5 Flash \& GPT-5.6 & 96.7\% \\
Gemini 2.5 Flash \& GPT-4.1 & 96.7\% \\
GPT-5.6 \& GPT-4.1         & 97.6\% \\
\bottomrule
\end{tabular}
\vspace{-6pt}
\end{wraptable}

\textbf{Agreement between different VLMs.} 
By default we use Gemini 2.5 Flash to answer the evaluation questions, mainly because of the low cost. %
Here we replace Gemini 2.5 Flash with GPT-5.6 and GPT-4.1 on the same human-study questions and keep the deterministic scoring part unchanged. The pairwise agreement between the three VLM judges is consistently greater than 96\%, as shown in Table~\ref{tab:reader_consistency}. It suggests that our evaluation protocol is robust with the current judge. %

\section{Conclusion}
\label{sec:Conclusion}
We introduce GenOlympiad, a challenging set of generation tasks covering nine capability axes to find where frontier T2I models fail. Each prompt has an evaluation program of easy and hard questions with 2.2 hard questions on average. We propose a grounding method for such hard question, where we draw helping lines on the image in an iterative way. We show that this method improves agreement with human judgments.%
We benchmark many strong models including open-source and close-source T2I models and show that the best T2I model, Nano Banana Pro, only has a strict score of 23.9. While coding models have higher strict score of 40.5, they suffer from poorer aesthetics. We also derive interesting insights, \emph{e.g.}, how a model family improves with their newer releases, prompt enhancer is useful, and GenOlympiad scores strongly correlate with human Elo scores.

\textbf{Limitation.} 
GenOlympiad focuses on prompts that current frontier models find difficult, so the scores do not represent their performance on easy prompts. Moreover, GenOlympiad offers limited signal for non-frontier T2I models. Some prompts in GenOlympiad are too long to fit into the context of open-weight models, so for some open-source models, around 20 tasks do not have image generated or scored.%
Some scores are also withheld because the VLM judge does not correctly ground evidence. Finally, our evaluation is cost more requiring 59.5 VLM reads per image on average.%
\subsection*{AI use statement}

In this work, we used generative AI tools to assist with generating synthetic benchmark tasks, developing the conceptual framework of the nine capability axes, designing and providing feedback on experiments, cleaning and reformatting the benchmark dataset, implementing evaluation programs, translating and summarizing prompts used while constructing the benchmark, as described in Appendix~\ref{sec:prompt}, evaluating generated images as part of the scoring pipeline, and providing preliminary interpretation of intermediate experimental results.
We have not used generative AI tools for any other tasks with required disclosure, and tasks involving formulating mathematical claims, providing critical ingredients for proving mathematical claims, assisting in the writing of proofs, proposing or refining hypotheses, and supporting qualitative and thematic data analysis are not applicable to this work.
Additionally, we used generative AI tools for brainstorming, sourcing and searching for information, summarizing and analyzing existing literature. 
AI-assisted benchmark tasks and evaluation programs were checked through manual spot checks and additional AI-based verification. The evaluation pipeline was separately validated against human annotations as described in the paper.
We take responsibility for the final content of this work, including text, claims or artifacts produced with the aid of generative AI.

\subsection*{Ethics statement}

Human annotations used in this work were collected from 41 participants recruited through Prolific. Participants were compensated at a rate of £13.20 per hour. We did not request or collect participants' names, email addresses, or other directly identifying information, nor did the study require participants to provide sensitive personal information.

\newpage

\bibliography{iclr2027_conference}

\begin{thebibliography}{52}
\providecommand{\natexlab}[1]{#1}
\providecommand{\url}[1]{\texttt{#1}}
\expandafter\ifx\csname urlstyle\endcsname\relax
  \providecommand{\doi}[1]{doi: #1}\else
  \providecommand{\doi}{doi: \begingroup \urlstyle{rm}\Url}\fi

\bibitem[Barbu et~al.(2019)Barbu, Mayo, Alverio, Luo, Wang, Gutfreund,
  Tenenbaum, and Katz]{objectnet}
Andrei Barbu, David Mayo, Julian Alverio, William Luo, Christopher Wang, Dan
  Gutfreund, Josh Tenenbaum, and Boris Katz.
\newblock Objectnet: {A} large-scale bias-controlled dataset for pushing the
  limits of object recognition models.
\newblock In Hanna~M. Wallach, Hugo Larochelle, Alina Beygelzimer, Florence
  d'Alch{\'{e}}{-}Buc, Emily~B. Fox, and Roman Garnett (eds.), \emph{Advances
  in Neural Information Processing Systems 32: Annual Conference on Neural
  Information Processing Systems 2019, NeurIPS 2019, December 8-14, 2019,
  Vancouver, BC, Canada}, pp.\  9448--9458, 2019.
\newblock URL
  \url{https://proceedings.neurips.cc/paper/2019/hash/97af07a14cacba681feacf3012730892-Abstract.html}.

\bibitem[{Black Forest Labs}(2024)]{flux1}
{Black Forest Labs}.
\newblock {FLUX.1}, 2024.
\newblock URL \url{https://github.com/black-forest-labs/flux}.
\newblock Accessed: 2026-09-14.

\bibitem[{Black Forest Labs}(2025)]{flux2}
{Black Forest Labs}.
\newblock {FLUX.2}, 2025.
\newblock URL \url{https://bfl.ai/blog/flux-2}.
\newblock Accessed: 2026-09-14.

\bibitem[{ByteDance Seed}(2025)]{seedream45}
{ByteDance Seed}.
\newblock {Seedream 4.5}, 2025.
\newblock URL \url{https://seed.bytedance.com/en/seedream4_5}.
\newblock Accessed: 2026-07-21.

\bibitem[{ByteDance Seed}(2026)]{seedream5}
{ByteDance Seed}.
\newblock {Seedream 5.0 Pro}, 2026.
\newblock URL \url{https://seed.bytedance.com/en/seedream5_0_pro}.
\newblock Accessed: 2026-09-10.

\bibitem[{Center for AI Safety} et~al.(2026){Center for AI Safety}, {Scale AI},
  and {HLE Contributors Consortium}]{hle}
{Center for AI Safety}, {Scale AI}, and {HLE Contributors Consortium}.
\newblock A benchmark of expert-level academic questions to assess {AI}
  capabilities.
\newblock \emph{Nature}, 649:\penalty0 1139--1146, 2026.
\newblock \doi{10.1038/s41586-025-09962-4}.
\newblock URL \url{https://arxiv.org/abs/2501.14249}.

\bibitem[Chang et~al.(2025)Chang, Fang, Xing, Wu, Cheng, Wang, Zeng, Yu, and
  Chen]{oneigbench}
Jingjing Chang, Yixiao Fang, Peng Xing, Shuhan Wu, Wei Cheng, Rui Wang,
  Xianfang Zeng, Gang Yu, and Hai{-}Bao Chen.
\newblock Oneig-bench: Omni-dimensional nuanced evaluation for image
  generation.
\newblock In Danielle Belgrave, Cheng Zhang, Laura~N. Montoya, Hsuan{-}Tien
  Lin, Razvan Pascanu, Piotr Koniusz, Marzyeh Ghassemi, Nancy Chen, Iv{\'{a}}n
  Vladimir~Meza Ru{\'{\i}}z, and Arturo Loaiza{-}Bonilla (eds.), \emph{Advances
  in Neural Information Processing Systems 38: Annual Conference on Neural
  Information Processing Systems 2025, NeurIPS 2025, San Diego, CA, USA,
  December 2-7, 2025 / Mexico City, Mexico, November 30 - December 5, 2025},
  2025.
\newblock URL
  \url{http://papers.nips.cc/paper_files/paper/2025/hash/e9e9e5428189a3e49479547ef917e88d-Abstract-Datasets_and_Benchmarks_Track.html}.

\bibitem[Chen et~al.(2025)Chen, Lin, Xu, Shen, Yao, Rimchala, Zhang, and
  Huang]{r2ibench}
Kaijie Chen, Zihao Lin, Zhiyang Xu, Ying Shen, Yuguang Yao, Joy Rimchala,
  Jiaxin Zhang, and Lifu Huang.
\newblock R2i-bench: Benchmarking reasoning-driven text-to-image generation.
\newblock In \emph{Proceedings of the 2025 Conference on Empirical Methods in
  Natural Language Processing}, pp.\  12595--12630, 2025.

\bibitem[Chen et~al.(2024)Chen, Zhang, Weng, Pan, and Lan]{PEStrongModel}
Zijie Chen, Lichao Zhang, Fangsheng Weng, Lili Pan, and Zhenzhong Lan.
\newblock Tailored visions: Enhancing text-to-image generation with
  personalized prompt rewriting.
\newblock In \emph{{IEEE/CVF} Conference on Computer Vision and Pattern
  Recognition, {CVPR} 2024, Seattle, WA, USA, June 16-22, 2024}, pp.\
  7727--7736. {IEEE}, 2024.
\newblock \doi{10.1109/CVPR52733.2024.00738}.
\newblock URL \url{https://doi.org/10.1109/CVPR52733.2024.00738}.

\bibitem[Cho et~al.(2024)Cho, Hu, Baldridge, Garg, Anderson, Krishna, Bansal,
  Pont-Tuset, and Wang]{dsg}
Jaemin Cho, Yushi Hu, Jason Baldridge, Roopal Garg, Peter Anderson, Ranjay
  Krishna, Mohit Bansal, Jordi Pont-Tuset, and Su~Wang.
\newblock Davidsonian scene graph: Improving reliability in fine-grained
  evaluation for text-to-image generation.
\newblock In \emph{International conference on learning representations}, 2024.

\bibitem[Chollet(2019)]{arc1}
Fran{\c{c}}ois Chollet.
\newblock On the measure of intelligence.
\newblock \emph{CoRR}, abs/1911.01547, 2019.
\newblock URL \url{http://arxiv.org/abs/1911.01547}.

\bibitem[Chollet et~al.(2025)Chollet, Knoop, Kamradt, Landers, and
  Pinkard]{arc2}
Fran{\c{c}}ois Chollet, Mike Knoop, Gregory Kamradt, Bryan Landers, and Henry
  Pinkard.
\newblock {ARC-AGI-2:} {A} new challenge for frontier {AI} reasoning systems.
\newblock \emph{CoRR}, abs/2505.11831, 2025.
\newblock \doi{10.48550/ARXIV.2505.11831}.
\newblock URL \url{https://doi.org/10.48550/arXiv.2505.11831}.

\bibitem[Esser et~al.(2024)Esser, Kulal, Blattmann, Entezari, M{\"{u}}ller,
  Saini, Levi, Lorenz, Sauer, Boesel, Podell, Dockhorn, English, and
  Rombach]{sd3}
Patrick Esser, Sumith Kulal, Andreas Blattmann, Rahim Entezari, Jonas
  M{\"{u}}ller, Harry Saini, Yam Levi, Dominik Lorenz, Axel Sauer, Frederic
  Boesel, Dustin Podell, Tim Dockhorn, Zion English, and Robin Rombach.
\newblock Scaling rectified flow transformers for high-resolution image
  synthesis.
\newblock In Ruslan Salakhutdinov, Zico Kolter, Katherine~A. Heller, Adrian
  Weller, Nuria Oliver, Jonathan Scarlett, and Felix Berkenkamp (eds.),
  \emph{Forty-first International Conference on Machine Learning, {ICML} 2024,
  Vienna, Austria, July 21-27, 2024}, volume 235 of \emph{Proceedings of
  Machine Learning Research}, pp.\  12606--12633. {PMLR} / OpenReview.net,
  2024.
\newblock URL \url{https://proceedings.mlr.press/v235/esser24a.html}.

\bibitem[Fang et~al.(2026)Fang, Yu, Duan, Huang, Bai, Cai, Wang, Liu, Liu, and
  Li]{prismbench}
Rongyao Fang, Aldrich Yu, Chengqi Duan, Linjiang Huang, Shuai Bai, Yuxuan Cai,
  Kun Wang, Si~Liu, Xihui Liu, and Hongsheng Li.
\newblock {FLUX}-reason-6m \& {PRISM}-bench: A million-scale text-to-image
  reasoning dataset and comprehensive benchmark.
\newblock In \emph{The Fourteenth International Conference on Learning
  Representations}, 2026.
\newblock URL \url{https://openreview.net/forum?id=cPzgZnpVbN}.

\bibitem[Ghosh et~al.(2023)Ghosh, Hajishirzi, and Schmidt]{geneval}
Dhruba Ghosh, Hannaneh Hajishirzi, and Ludwig Schmidt.
\newblock Geneval: An object-focused framework for evaluating text-to-image
  alignment.
\newblock In Alice Oh, Tristan Naumann, Amir Globerson, Kate Saenko, Moritz
  Hardt, and Sergey Levine (eds.), \emph{Advances in Neural Information
  Processing Systems 36: Annual Conference on Neural Information Processing
  Systems 2023, NeurIPS 2023, New Orleans, LA, USA, December 10 - 16, 2023},
  2023.
\newblock URL
  \url{https://papers.nips.cc/paper_files/paper/2023/hash/a3bf71c7c63f0c3bcb7ff67c67b1e7b1-Abstract-Datasets_and_Benchmarks.html}.

\bibitem[Glazer et~al.(2024)Glazer, Erdil, Besiroglu, Chicharro, Chen, Gunning,
  Olsson, Denain, Ho, de~Oliveira~Santos, J{\"{a}}rviniemi, Barnett, Sandler,
  Vrzala, Sevilla, Ren, Pratt, Levine, Barkley, Stewart, Grechuk, Grechuk,
  Enugandla, and Wildon]{frontiermath}
Elliot Glazer, Ege Erdil, Tamay Besiroglu, Diego Chicharro, Evan Chen, Alex
  Gunning, Caroline~Falkman Olsson, Jean{-}Stanislas Denain, Anson Ho, Emily
  de~Oliveira~Santos, Olli J{\"{a}}rviniemi, Matthew Barnett, Robert Sandler,
  Matej Vrzala, Jaime Sevilla, Qiuyu Ren, Elizabeth Pratt, Lionel Levine, Grant
  Barkley, Natalie Stewart, Bogdan Grechuk, Tetiana Grechuk, Shreepranav~Varma
  Enugandla, and Mark Wildon.
\newblock Frontiermath: {A} benchmark for evaluating advanced mathematical
  reasoning in {AI}.
\newblock \emph{CoRR}, abs/2411.04872, 2024.
\newblock \doi{10.48550/ARXIV.2411.04872}.
\newblock URL \url{https://doi.org/10.48550/arXiv.2411.04872}.

\bibitem[{Google}(2025{\natexlab{a}})]{nanobanana}
{Google}.
\newblock {Nano Banana (Gemini 2.5 Flash Image)}, 2025{\natexlab{a}}.
\newblock URL
  \url{https://developers.googleblog.com/en/introducing-gemini-2-5-flash-image/}.
\newblock Accessed: 2026-09-01.

\bibitem[{Google}(2025{\natexlab{b}})]{nanobananapro}
{Google}.
\newblock {Nano Banana Pro (Gemini 3 Pro Image)}, 2025{\natexlab{b}}.
\newblock URL
  \url{https://blog.google/innovation-and-ai/products/nano-banana-pro/}.
\newblock Accessed: 2026-09-14.

\bibitem[{Google}(2026)]{nanobanana2}
{Google}.
\newblock {Nano Banana 2 (Gemini 3.1 Flash Image)}, 2026.
\newblock URL
  \url{https://blog.google/innovation-and-ai/technology/ai/nano-banana-2/}.
\newblock Accessed: 2026-07-21.

\bibitem[Hessel et~al.(2021)Hessel, Holtzman, Forbes, Le~Bras, and
  Choi]{clipscore}
Jack Hessel, Ari Holtzman, Maxwell Forbes, Ronan Le~Bras, and Yejin Choi.
\newblock Clipscore: A reference-free evaluation metric for image captioning.
\newblock In \emph{Proceedings of the 2021 conference on empirical methods in
  natural language processing}, pp.\  7514--7528, 2021.

\bibitem[Heusel et~al.(2017)Heusel, Ramsauer, Unterthiner, Nessler, and
  Hochreiter]{fid}
Martin Heusel, Hubert Ramsauer, Thomas Unterthiner, Bernhard Nessler, and Sepp
  Hochreiter.
\newblock Gans trained by a two time-scale update rule converge to a local nash
  equilibrium.
\newblock In Isabelle Guyon, Ulrike von Luxburg, Samy Bengio, Hanna~M. Wallach,
  Rob Fergus, S.~V.~N. Vishwanathan, and Roman Garnett (eds.), \emph{Advances
  in Neural Information Processing Systems 30: Annual Conference on Neural
  Information Processing Systems 2017, December 4-9, 2017, Long Beach, CA,
  {USA}}, pp.\  6626--6637, 2017.
\newblock URL
  \url{https://proceedings.neurips.cc/paper/2017/hash/8a1d694707eb0fefe65871369074926d-Abstract.html}.

\bibitem[Hu et~al.(2024)Hu, Wang, Fang, Fu, Cheng, and Yu]{ella}
Xiwei Hu, Rui Wang, Yixiao Fang, Bin Fu, Pei Cheng, and Gang Yu.
\newblock {ELLA:} equip diffusion models with {LLM} for enhanced semantic
  alignment.
\newblock \emph{CoRR}, abs/2403.05135, 2024.
\newblock \doi{10.48550/ARXIV.2403.05135}.
\newblock URL \url{https://doi.org/10.48550/arXiv.2403.05135}.

\bibitem[Hu et~al.(2023)Hu, Liu, Kasai, Wang, Ostendorf, Krishna, and
  Smith]{tifa}
Yushi Hu, Benlin Liu, Jungo Kasai, Yizhong Wang, Mari Ostendorf, Ranjay
  Krishna, and Noah~A Smith.
\newblock Tifa: Accurate and interpretable text-to-image faithfulness
  evaluation with question answering.
\newblock In \emph{2023 ieee/cvf international conference on computer vision
  (iccv)}, pp.\  20349--20360. IEEE, 2023.

\bibitem[Huang et~al.(2025)Huang, Duan, Sun, Xie, Li, and Liu]{compbench++}
Kaiyi Huang, Chengqi Duan, Kaiyue Sun, Enze Xie, Zhenguo Li, and Xihui Liu.
\newblock T2i-compbench++: An enhanced and comprehensive benchmark for
  compositional text-to-image generation.
\newblock \emph{{IEEE} Trans. Pattern Anal. Mach. Intell.}, 47\penalty0
  (5):\penalty0 3563--3579, 2025.
\newblock \doi{10.1109/TPAMI.2025.3531907}.
\newblock URL \url{https://doi.org/10.1109/TPAMI.2025.3531907}.

\bibitem[Jimenez et~al.(2024)Jimenez, Yang, Wettig, Yao, Pei, Press, and
  Narasimhan]{swebench}
Carlos~E. Jimenez, John Yang, Alexander Wettig, Shunyu Yao, Kexin Pei, Ofir
  Press, and Karthik~R. Narasimhan.
\newblock Swe-bench: Can language models resolve real-world github issues?
\newblock In \emph{The Twelfth International Conference on Learning
  Representations, {ICLR} 2024, Vienna, Austria, May 7-11, 2024}.
  OpenReview.net, 2024.
\newblock URL \url{https://openreview.net/forum?id=VTF8yNQM66}.

\bibitem[Kamath et~al.(2025)Kamath, Chang, Krishna, Zettlemoyer, Hu, and
  Ghazvininejad]{geneval2}
Amita Kamath, Kai{-}Wei Chang, Ranjay Krishna, Luke Zettlemoyer, Yushi Hu, and
  Marjan Ghazvininejad.
\newblock Geneval 2: Addressing benchmark drift in text-to-image evaluation.
\newblock \emph{CoRR}, abs/2512.16853, 2025.
\newblock \doi{10.48550/ARXIV.2512.16853}.
\newblock URL \url{https://doi.org/10.48550/arXiv.2512.16853}.

\bibitem[Li et~al.(2024)Li, Lin, Pathak, Li, Fei, Wu, Ling, Xia, Zhang, Neubig,
  and Ramanan]{genaibench}
Baiqi Li, Zhiqiu Lin, Deepak Pathak, Jiayao Li, Yixin Fei, Kewen Wu, Tiffany
  Ling, Xide Xia, Pengchuan Zhang, Graham Neubig, and Deva Ramanan.
\newblock Genai-bench: Evaluating and improving compositional text-to-visual
  generation.
\newblock \emph{CoRR}, abs/2406.13743, 2024.
\newblock \doi{10.48550/ARXIV.2406.13743}.
\newblock URL \url{https://doi.org/10.48550/arXiv.2406.13743}.

\bibitem[Li et~al.(2026)Li, Hu, Qiao, Ba, Hong, Shen, Wang, Zhou, Kang, Shang,
  He, Wang, Li, Li, Zhang, Gao, Yan, Jiang, Tang, Yin, Wu, Xu, Chen, Chen, Shu,
  Zhang, Chen, Xu, Zhang, Wang, Liu, Zhou, Shi, Wang, Zhao, Wei, Qu, and
  Wu]{qwenbench}
Niantong Li, Guangzheng Hu, Weixu Qiao, Ying Ba, Qichen Hong, Shijun Shen,
  Jinlin Wang, Fan Zhou, Jianye Kang, Xin Shang, Ziyi He, Wei Wang, Dalin Li,
  Jiahao Li, Jie Zhang, Kaiyuan Gao, Kun Yan, Lihan Jiang, Ningyuan Tang,
  Shengming Yin, Tianhe Wu, Xiao Xu, Xiaoyue Chen, Yuxiang Chen, Yan Shu,
  Yanran Zhang, Yilei Chen, Yixian Xu, Zekai Zhang, Zhendong Wang, Zihao Liu,
  Zikai Zhou, Hongzhu Shi, Yi~Wang, Bing Zhao, Hu~Wei, Lin Qu, and Chenfei Wu.
\newblock Qwen-image-bench: From generation to creation in text-to-image
  evaluation.
\newblock \emph{CoRR}, abs/2605.28091, 2026.
\newblock \doi{10.48550/ARXIV.2605.28091}.
\newblock URL \url{https://doi.org/10.48550/arXiv.2605.28091}.

\bibitem[Lin et~al.(2024)Lin, Pathak, Li, Li, Xia, Neubig, Zhang, and
  Ramanan]{vqascore}
Zhiqiu Lin, Deepak Pathak, Baiqi Li, Jiayao Li, Xide Xia, Graham Neubig,
  Pengchuan Zhang, and Deva Ramanan.
\newblock Evaluating text-to-visual generation with image-to-text generation.
\newblock In \emph{European Conference on Computer Vision}, pp.\  366--384.
  Springer, 2024.

\bibitem[Liu et~al.(2026)Liu, Shen, Zhang, Hsieh, Zhang, Xu, Chen, Li, Lu, Ma,
  et~al.]{mathgen}
Ruiyao Liu, Hui Shen, Ping Zhang, Yunta Hsieh, Yifan Zhang, Jing Xu, Sicheng
  Chen, Junchen Li, Jiawei Lu, Jianing Ma, et~al.
\newblock Mathgen: Revealing the illusion of mathematical competence through
  text-to-image generation.
\newblock \emph{arXiv preprint arXiv:2603.27959}, 2026.

\bibitem[Niu et~al.(2025)Niu, Ning, Zheng, Lin, Jin, Liao, Ning, Zhu, and
  Yuan]{wise}
Yuwei Niu, Munan Ning, Mengren Zheng, Bin Lin, Peng Jin, Jiaqi Liao, Kun{-}Peng
  Ning, Bin Zhu, and Li~Yuan.
\newblock {WISE:} {A} world knowledge-informed semantic evaluation for
  text-to-image generation.
\newblock \emph{CoRR}, abs/2503.07265, 2025.
\newblock \doi{10.48550/ARXIV.2503.07265}.
\newblock URL \url{https://doi.org/10.48550/arXiv.2503.07265}.

\bibitem[{OpenAI}(2025{\natexlab{a}})]{gpt41}
{OpenAI}.
\newblock {GPT-4.1}, 2025{\natexlab{a}}.
\newblock URL \url{https://openai.com/index/gpt-4-1/}.
\newblock Accessed: 2026-09-04.

\bibitem[{OpenAI}(2025{\natexlab{b}})]{gptimage1}
{OpenAI}.
\newblock {GPT-Image-1}, 2025{\natexlab{b}}.
\newblock URL \url{https://openai.com/index/introducing-4o-image-generation/}.
\newblock Accessed: 2026-09-01.

\bibitem[{OpenAI}(2025{\natexlab{c}})]{gptimage15}
{OpenAI}.
\newblock {GPT-Image-1.5}, 2025{\natexlab{c}}.
\newblock URL \url{https://openai.com/index/new-chatgpt-images-is-here/}.
\newblock Accessed: 2026-07-27.

\bibitem[{OpenAI}(2026{\natexlab{a}})]{gpt56}
{OpenAI}.
\newblock {GPT-5.6 Sol}, 2026{\natexlab{a}}.
\newblock URL \url{https://openai.com/index/gpt-5-6/}.
\newblock Accessed: 2026-09-15.

\bibitem[{OpenAI}(2026{\natexlab{b}})]{gpt6astra}
{OpenAI}.
\newblock {GPT-6 Astra}, 2026{\natexlab{b}}.
\newblock URL \url{https://openai.com/index/gpt-6-astra/}.
\newblock Accessed: 2026-09-14.

\bibitem[{OpenAI}(2026{\natexlab{c}})]{gptimage2}
{OpenAI}.
\newblock {GPT-Image-2}, 2026{\natexlab{c}}.
\newblock URL \url{https://openai.com/index/introducing-chatgpt-images-2-0/}.
\newblock Accessed: 2026-07-18.

\bibitem[{OpenAI}(2026{\natexlab{d}})]{gptimage25}
{OpenAI}.
\newblock {GPT-Image-2.5}, 2026{\natexlab{d}}.
\newblock URL \url{https://openai.com/index/introducing-chatgpt-images-2-5/}.
\newblock Accessed: 2026-09-09.

\bibitem[Paiss et~al.(2023)Paiss, Ephrat, Tov, Zada, Mosseri, Irani, and
  Dekel]{countbench}
Roni Paiss, Ariel Ephrat, Omer Tov, Shiran Zada, Inbar Mosseri, Michal Irani,
  and Tali Dekel.
\newblock Teaching {CLIP} to count to ten.
\newblock In \emph{{IEEE/CVF} International Conference on Computer Vision,
  {ICCV} 2023, Paris, France, October 1-6, 2023}, pp.\  3147--3157. {IEEE},
  2023.
\newblock \doi{10.1109/ICCV51070.2023.00294}.
\newblock URL \url{https://doi.org/10.1109/ICCV51070.2023.00294}.

\bibitem[Salimans et~al.(2016)Salimans, Goodfellow, Zaremba, Cheung, Radford,
  and Chen]{is}
Tim Salimans, Ian~J. Goodfellow, Wojciech Zaremba, Vicki Cheung, Alec Radford,
  and Xi~Chen.
\newblock Improved techniques for training gans.
\newblock In Daniel~D. Lee, Masashi Sugiyama, Ulrike von Luxburg, Isabelle
  Guyon, and Roman Garnett (eds.), \emph{Advances in Neural Information
  Processing Systems 29: Annual Conference on Neural Information Processing
  Systems 2016, December 5-10, 2016, Barcelona, Spain}, pp.\  2226--2234, 2016.
\newblock URL
  \url{https://proceedings.neurips.cc/paper/2016/hash/8a3363abe792db2d8761d6403605aeb7-Abstract.html}.

\bibitem[Sani et~al.(2026)Sani, Ku, Jamali, Sani, Khoshtab, Sun, Fazel, Tam,
  Chong, Chan, Tsang, Hsu, Lam, Ng, Chu, Mak, Wu, Wong, Ho, Ruan, Li, Fang,
  Yeh, Cheng, Nie, and Chen]{imagenworld}
Samin~Mahdizadeh Sani, Max Ku, Nima Jamali, Matina~Mahdizadeh Sani, Paria
  Khoshtab, Wei{-}Chieh Sun, Parnian Fazel, Zhi~Rui Tam, Thomas Chong, Edisy
  Kin~Wai Chan, Donald Wai~Tong Tsang, Chiao{-}Wei Hsu, Ting~Wai Lam,
  Ho~Yin~Sam Ng, Chiafeng Chu, Chak{-}Wing Mak, Keming Wu, Hiu~Tung Wong,
  Yik~Chun Ho, Chi Ruan, Zhuofeng Li, I{-}Sheng Fang, Shih{-}Ying Yeh, Ho~Kei
  Cheng, Ping Nie, and Wenhu Chen.
\newblock Imagenworld: Stress-testing image generation models with explainable
  human evaluation on open-ended real-world tasks.
\newblock \emph{CoRR}, abs/2603.27862, 2026.
\newblock \doi{10.48550/ARXIV.2603.27862}.
\newblock URL \url{https://doi.org/10.48550/arXiv.2603.27862}.

\bibitem[Singh \& Zheng(2023)Singh and Zheng]{dscore}
Jaskirat Singh and Liang Zheng.
\newblock Divide, evaluate, and refine: Evaluating and improving text-to-image
  alignment with iterative {VQA} feedback.
\newblock In Alice Oh, Tristan Naumann, Amir Globerson, Kate Saenko, Moritz
  Hardt, and Sergey Levine (eds.), \emph{Advances in Neural Information
  Processing Systems 36: Annual Conference on Neural Information Processing
  Systems 2023, NeurIPS 2023, New Orleans, LA, USA, December 10 - 16, 2023},
  2023.
\newblock URL
  \url{http://papers.nips.cc/paper_files/paper/2023/hash/dfd0bd56e8a6f82d1619f5d093d5f9ca-Abstract-Conference.html}.

\bibitem[{Stability AI}(2024)]{sd35}
{Stability AI}.
\newblock {Stable Diffusion 3.5}, 2024.
\newblock URL \url{https://github.com/Stability-AI/sd3.5}.
\newblock Accessed: 2026-09-06.

\bibitem[Sun et~al.(2026)Sun, Fang, Duan, Liu, Li, and Liu]{t2ireasonbench}
Kaiyue Sun, Rongyao Fang, Chengqi Duan, Xian Liu, Aoxue Li, and Xihui Liu.
\newblock T2i-reasonbench: Benchmarking reasoning-informed text-to-image
  generation.
\newblock In \emph{Findings of the Association for Computational Linguistics:
  ACL 2026}, pp.\  8919--8944, 2026.

\bibitem[Wang et~al.(2026)Wang, Xu, Xing, Cheng, Zhao, Li, Hang, Li, Tao, Wang,
  Li, Chen, Li, Wu, Deng, Gu, Wang, and Lu]{PE32B}
Linqing Wang, Zhiyong Xu, Ximing Xing, Yiji Cheng, Zhiyuan Zhao, Donghao Li,
  Tiankai Hang, Zhenxi Li, Jiale Tao, Qixun Wang, Ruihuang Li, Comi Chen, Xin
  Li, Mingrui Wu, Xinchi Deng, Shuyang Gu, Chunyu Wang, and Qinglin Lu.
\newblock Promptenhancer: Taming your rewriter for text-to-image generation via
  fine-grained reward.
\newblock In \emph{Proceedings of the IEEE/CVF Conference on Computer Vision
  and Pattern Recognition (CVPR)}, pp.\  14895--14904, June 2026.

\bibitem[Wang et~al.(2025)Wang, Yin, Zhao, Tian, Qiao, Wang, Dai, and
  Luo]{genexam}
Zhaokai Wang, Penghao Yin, Xiangyu Zhao, Changyao Tian, Yu~Qiao, Wenhai Wang,
  Jifeng Dai, and Gen Luo.
\newblock Genexam: {A} multidisciplinary text-to-image exam.
\newblock \emph{CoRR}, abs/2509.14232, 2025.
\newblock \doi{10.48550/ARXIV.2509.14232}.
\newblock URL \url{https://doi.org/10.48550/arXiv.2509.14232}.

\bibitem[Wei et~al.(2025)Wei, Zhang, Wang, Wei, Guo, and Zhang]{tiifbench}
Xinyu Wei, Jinrui Zhang, Zeqing Wang, Hongyang Wei, Zhen Guo, and Lei Zhang.
\newblock Tiif-bench: How does your {T2I} model follow your instructions?
\newblock \emph{CoRR}, abs/2506.02161, 2025.
\newblock \doi{10.48550/ARXIV.2506.02161}.
\newblock URL \url{https://doi.org/10.48550/arXiv.2506.02161}.

\bibitem[Wu et~al.(2025)Wu, Li, Zhou, Lin, Gao, Yan, Yin, Bai, Xu, Chen, Chen,
  Tang, Zhang, Wang, Yang, Yu, Cheng, Liu, Li, Zhang, Meng, Wei, Ni, Chen, Cao,
  Peng, Qu, Wu, Wang, Yu, Wen, Feng, Xu, Wang, Zhang, Zhu, Wu, Cai, and
  Liu]{qwenimage}
Chenfei Wu, Jiahao Li, Jingren Zhou, Junyang Lin, Kaiyuan Gao, Kun Yan,
  Shengming Yin, Shuai Bai, Xiao Xu, Yilei Chen, Yuxiang Chen, Zecheng Tang,
  Zekai Zhang, Zhengyi Wang, An~Yang, Bowen Yu, Chen Cheng, Dayiheng Liu,
  Deqing Li, Hang Zhang, Hao Meng, Hu~Wei, Jingyuan Ni, Kai Chen, Kuan Cao,
  Liang Peng, Lin Qu, Minggang Wu, Peng Wang, Shuting Yu, Tingkun Wen, Wensen
  Feng, Xiaoxiao Xu, Yi~Wang, Yichang Zhang, Yongqiang Zhu, Yujia Wu, Yuxuan
  Cai, and Zenan Liu.
\newblock Qwen-image technical report.
\newblock \emph{CoRR}, abs/2508.02324, 2025.
\newblock \doi{10.48550/ARXIV.2508.02324}.
\newblock URL \url{https://doi.org/10.48550/arXiv.2508.02324}.

\bibitem[Wu et~al.(2024)Wu, Yu, Huang, Russakovsky, and Arora]{conceptmix}
Xindi Wu, Dingli Yu, Yangsibo Huang, Olga Russakovsky, and Sanjeev Arora.
\newblock Conceptmix: {A} compositional image generation benchmark with
  controllable difficulty.
\newblock In Amir Globersons, Lester Mackey, Danielle Belgrave, Angela Fan,
  Ulrich Paquet, Jakub~M. Tomczak, and Cheng Zhang (eds.), \emph{Advances in
  Neural Information Processing Systems 37: Annual Conference on Neural
  Information Processing Systems 2024, NeurIPS 2024, Vancouver, BC, Canada,
  December 10 - 15, 2024}, 2024.
\newblock URL
  \url{http://papers.nips.cc/paper_files/paper/2024/hash/9c3563bbeb2ad7f3b3b8ed0fcd3b440f-Abstract-Datasets_and_Benchmarks_Track.html}.

\bibitem[Xie et~al.(2025)Xie, Chen, Chen, Cai, Tang, Lin, Zhang, Li, Zhu, Lu,
  and Han]{sana}
Enze Xie, Junsong Chen, Junyu Chen, Han Cai, Haotian Tang, Yujun Lin, Zhekai
  Zhang, Muyang Li, Ligeng Zhu, Yao Lu, and Song Han.
\newblock {SANA:} efficient high-resolution text-to-image synthesis with linear
  diffusion transformers.
\newblock In \emph{The Thirteenth International Conference on Learning
  Representations, {ICLR} 2025, Singapore, April 24-28, 2025}. OpenReview.net,
  2025.
\newblock URL \url{https://openreview.net/forum?id=N8Oj1XhtYZ}.

\bibitem[Yang et~al.(2025)Yang, Zhao, Wan, and Ye]{thinkingwithimage}
Wenxi Yang, Yuzhong Zhao, Fang Wan, and Qixiang Ye.
\newblock Thinking with images via self-calling agent.
\newblock \emph{CoRR}, abs/2512.08511, 2025.
\newblock \doi{10.48550/ARXIV.2512.08511}.
\newblock URL \url{https://doi.org/10.48550/arXiv.2512.08511}.

\bibitem[Ye et~al.(2025)Ye, Jiang, Wang, Zhu, Hu, Huang, He, Yan, Yu, Li, He,
  and Li]{echo4o}
Junyan Ye, Dongzhi Jiang, Zihao Wang, Leqi Zhu, Zhenghao Hu, Zilong Huang, Jun
  He, Zhiyuan Yan, Jinghua Yu, Hongsheng Li, Conghui He, and Weijia Li.
\newblock Echo-4o: Harnessing the power of gpt-4o synthetic images for improved
  image generation.
\newblock \emph{CoRR}, abs/2508.09987, 2025.
\newblock \doi{10.48550/ARXIV.2508.09987}.
\newblock URL \url{https://doi.org/10.48550/arXiv.2508.09987}.

\end{thebibliography}
\bibliographystyle{iclr2027_conference}

\appendix
\newpage
\section{Appendix}

\subsection{More Examples and Benchmark Statistic}
\begin{figure}[t]
    \centering
    \includegraphics[width=\textwidth]{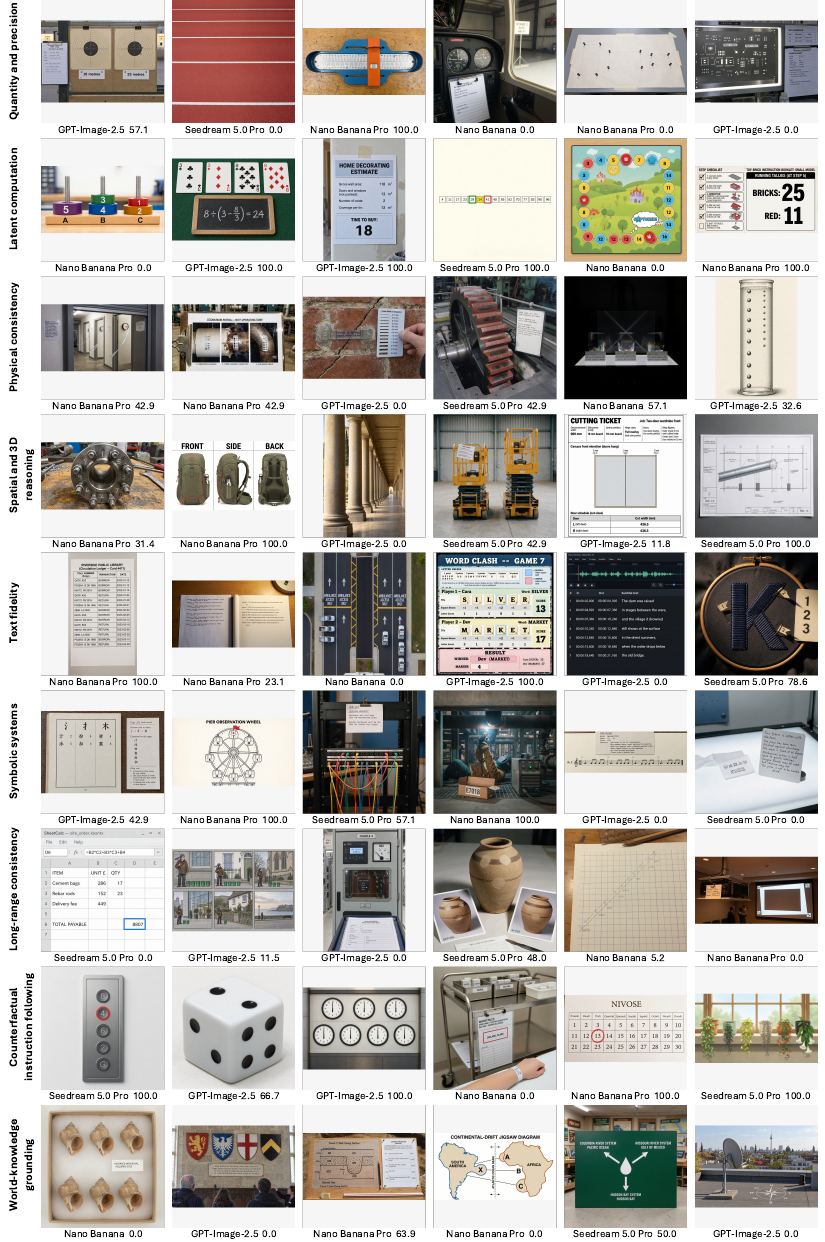}
    \caption{
        More example images for each capability axis.
    }
    \label{fig:wall}
\end{figure}

\paragraph{More Examples.} More example images of GenOlympiad are shown in Fig.\ref{fig:wall}. We have many different topic for each capability axis. The task topics in each axis can be seen in  Fig.~\ref{fig:axistopic}. Each prompt passed our task selection protocol described in Appendix~\ref{sec:howwebuild}. When designing tasks, we tried to limit simple changes such as changing `counting apples' to `counting footballs'. We also tried to make the mechanism of the tasks as diverse as possible. For example, our counting tasks include counting absent seats while other seats occupied by people, counting caps of one color among many mixed-color caps in Fig.\ref{fig:eval}, and counting how many beads of the same color are next to each other on a necklace. 
\begin{figure}[h]
    \centering
    \includegraphics[width=\textwidth]{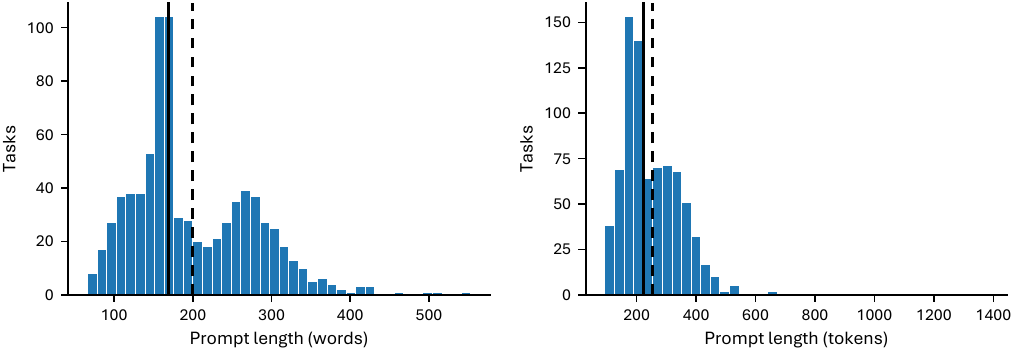}
    \caption{
        Prompt length distribution, \textbf{Left:} word count distribution, \textbf{Right:} Token distribution using \texttt{o200k\_base}. \textbf{Solid line}: median, \textbf{Dashed line}: mean.
    }
    \label{fig:prompt_length}
\end{figure}
\paragraph{Prompt Length.} We show the prompt length distribution of GenOlympiad in Fig.~\ref{fig:prompt_length}. The prompts contain 199 words in average, with a median of 169 words. We use \texttt{o200k\_base} to count tokens, each prompt has 254 tokens on average, and the median is 222 tokens.
Most prompts are shorter than 512 tokens.

\paragraph{Is GenOlympiad stably hard?} We benchmark GPT-Image-2.5 sunburst twice, the performance between the 2 attempts kept stable. Result is shown in Tab.~\ref{tab:second_try}. The performance of the 2 attempts is stable: the strict score is 14.9 and 14.2 and the relaxed score is 29.6 and 29.3. Besides, there are 702 tasks that get the same score in both attempts. Though there are some different in some axis in the 2 attempts, GenOlympiad is overall stably hard.

\begin{table}[h]
\centering
\caption{
GenOlympiad results for GPT-Image-2.5 Sunburst for its first and second attempts of GenOlympiad across nine capability axes. Best result in each column is highlighted in \textbf{bold}. C-Fact.: counterfactual instruction following; Symbol.: symbolic systems; World.: world-knowledge grounding; Phys.: physical consistency; Quant.: quantity and precision; Spatial.: spatial and 3D reasoning; Latent.: latent computation; Long.: long-range consistency. Overall is the mean score over all scored tasks; Str. and Rel. are strict and relaxed scores.
}
\label{tab:second_try}

\scriptsize
\setlength{\tabcolsep}{0pt}
\renewcommand{\arraystretch}{1.1}
\begin{tabular*}{\textwidth}{
@{\extracolsep{\fill}}
l@{\hspace{1.5pt}}|
cc cc cc cc cc cc cc cc cc
@{\hspace{5pt}}|
cc
@{}
}
\toprule

\textbf{Model}
& \multicolumn{2}{c}{\textbf{Text}}
& \multicolumn{2}{c}{\textbf{C-Fact.}}
& \multicolumn{2}{c}{\textbf{Symbol.}}
& \multicolumn{2}{c}{\textbf{World.}}
& \multicolumn{2}{c}{\textbf{Phys.}}
& \multicolumn{2}{c}{\textbf{Quant.}}
& \multicolumn{2}{c}{\textbf{Spatial.}}
& \multicolumn{2}{c}{\textbf{Latent.}}
& \multicolumn{2}{c}{\textbf{Long.}}
& \multicolumn{2}{c}{\textbf{Overall}}
\\

&
\textbf{Str.} & \textbf{Rel.}
& \textbf{Str.} & \textbf{Rel.}
& \textbf{Str.} & \textbf{Rel.}
& \textbf{Str.} & \textbf{Rel.}
& \textbf{Str.} & \textbf{Rel.}
& \textbf{Str.} & \textbf{Rel.}
& \textbf{Str.} & \textbf{Rel.}
& \textbf{Str.} & \textbf{Rel.}
& \textbf{Str.} & \textbf{Rel.}
& \textbf{Str.} & \textbf{Rel.}
\\
\midrule

GPT-Image-2.5
& \textbf{30.4} & \textbf{41.8}
& \textbf{23.0} & \textbf{36.4}
& \textbf{20.5} & \textbf{34.8}
& 6.8 & 24.1 
& 8.1 & 23.7 
& \textbf{13.2} & 25.0 
& \textbf{18.4} & \textbf{34.6} 
& 6.5 & 21.9 
& 9.8 & 25.9 
& \textbf{14.9} & \textbf{29.6} \\

GPT-Image-2.5
& 25.6 & 40.3 
& 18.4 & 35.1
& 13.5 & 27.8 
& \textbf{11.5} & \textbf{25.4}
& \textbf{10.8} & \textbf{27.9} 
& 12.7 & \textbf{26.2}
& 16.5 & 31.1 
& \textbf{10.4} & \textbf{24.2} 
& \textbf{9.9} & \textbf{26.3} 
& 14.2 & 29.3 \\

\bottomrule
\end{tabular*}
\end{table}

\begin{figure}[t]
    \centering

    \begin{minipage}[t]{0.49\textwidth}
        \centering
        \includegraphics[width=\linewidth]{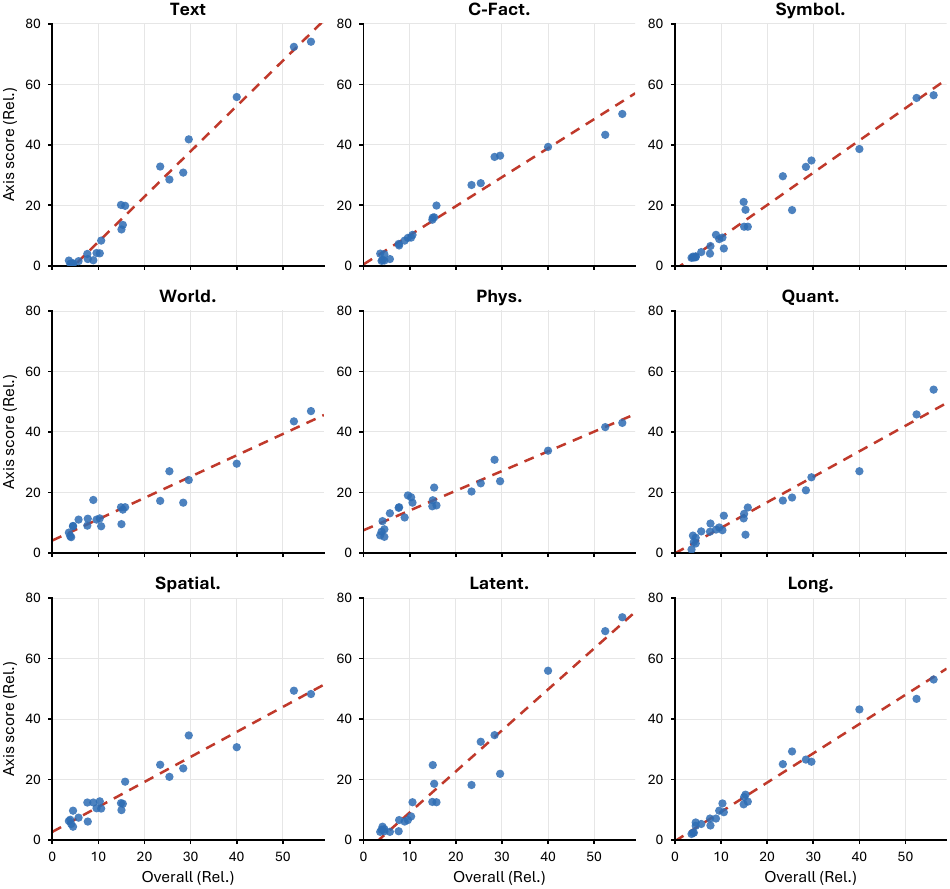}
        \captionof{figure}{
        Linear regression for calculating partial correlation.
        }
        \label{fig:axisandoverall}
    \end{minipage}
    \hfill
    \begin{minipage}[t]{0.49\textwidth}
        \centering
        \includegraphics[width=\linewidth]{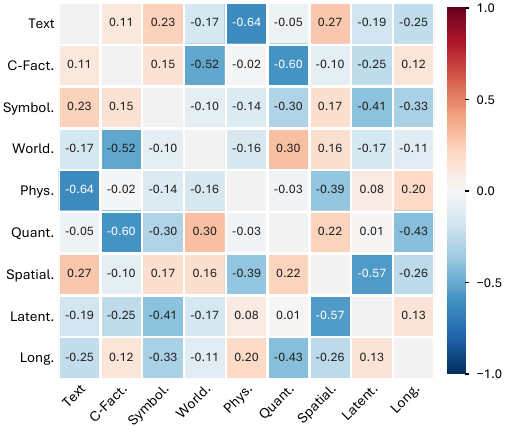}
        \captionof{figure}{
        Spearman correlation matrix across capability axes.
        }
        \label{fig:corr_res}
    \end{minipage}

\end{figure}

\paragraph{Axis Correlation.} Although we try to design different axes to measure different capabilities of T2I models, it is unavoidable that there are still many overlaps between prompts not only in the same axes, but also in different axes. For example, \textbf{physical consistency} contains many tasks that is the combination of \textbf{latent computation} and \textbf{world-knowledge grounding}. \textbf{Text fidelity} is also unavoidable to have some tasks that requires \textbf{quantitative precision} \textit{e.g.}, there should be 12 lines of text end with -ation. So we further measure the correlation between axes. The correlation matrix can be seen in Fig.~\ref{fig:corr_res}. Most pairs of axes have low correlation, while several pairs even show negative correlation.

\paragraph{The method we use to draw the correlation matrix.} We take a classic statistics method called \textbf{partial correlation} to calculate the correlation between each axis. We do not directly compare the scores of two axes, because stronger models usually get higher scores on many axes at the same time. So we take methods to mitigate the bias from models' capability by using linear regression. For each axis, we first predict what score a model should get based on its overall performance using linear regression shown in Fig.~\ref{fig:axisandoverall}. We then compare whether the model performs better or worse than this expected score on different axes. For example, if models that perform better than expected on \textbf{latent computation} can perform either better or worse than expected on \textbf{physical consistency}, then these two axes have a low spearman correlation. 

\begin{figure}[t]
    \centering
    \includegraphics[width=\textwidth]{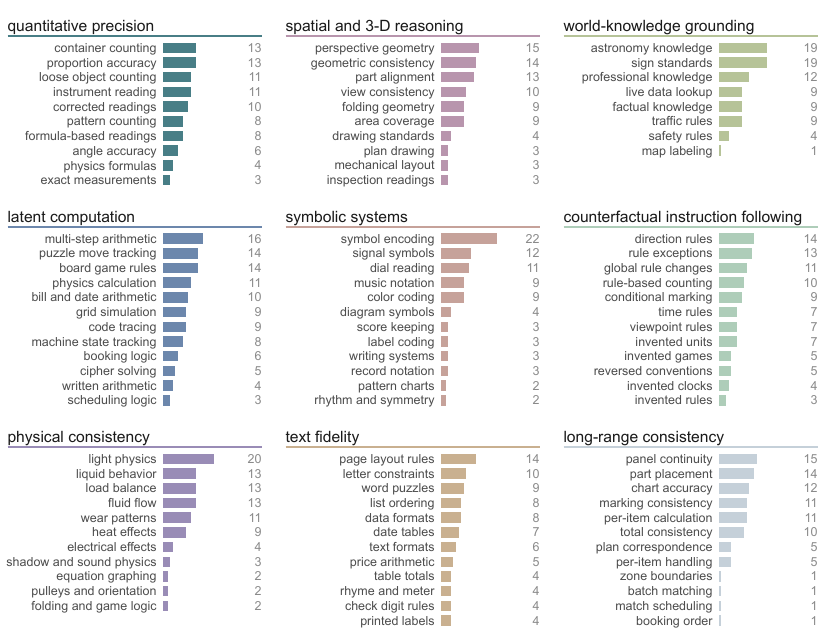}
    \caption{
        Different Topics for each axis.
    }
    \label{fig:axistopic}
\end{figure}

\subsection{Additional Analysis of Evaluator Agreement}
\textbf{The distribution of pass/fail of human agreement} 
To ensure that our high agreement between the automatic evaluation protocol and human judgments is not because the imbalanced distribution of pass and fail, we further report the confusion matrix in Tab.~\ref{tab:human_pass_fail}. According to the confusion matrix, our agreement is not mainly affected by pass or fail.

\begin{table}[t]
\centering
\small
\caption{Pass/fail distribution between human judgments and our automatic evaluator.}
\label{tab:human_pass_fail}
\begin{tabular}{lccc}
\toprule
 & \textbf{Auto Pass} & \textbf{Auto Fail} & \textbf{Total} \\
\midrule
Human Pass & 174 & 13 & 187 \\
Human Fail & 4 & 174 & 178 \\
\midrule
Total & 178 & 187 & 365 \\
\bottomrule
\end{tabular}
\end{table}

\subsection{How do we build GenOlympiad?}
\label{sec:howwebuild}
We designed an automated agentic task generate protocol to help us create high quality tasks. We provide simplified prompts of the protocol in Appendix~\ref{sec:prompt}

This protocol is made up of several parts. We first build about 1600 task descriptions by using a brainstorming agent powered by Claude Opus 5. We provided many positive examples and negative examples designed by human, then use the prompt in Appendix~\ref{sec:prompt} to help us conduct large scale brainstorming. Then we do the following check, filtering, then build tasks to get the final 801 tasks.
\begin{itemize}[leftmargin=*]
    \item \textbf{Detailed definition of Each Axis}. The detailed explanation about each capability axis with examples to guide Claude Opus 5 how to generate tasks.
    \item  \textbf{Few-shot Example for high-quality tasks}. We manually designed several origin T2I prompts for each capability as few-shot examples for agent. Each prompt reveals a typical fail of GPT-Image-2. 
    \item \textbf{Example to build evaluation program}. We designed a modular evaluation framework consists of 43 operators, each operator contains an atomic operation \emph{e.g.}, Cut the image by using coordination we provide. 
    \item \textbf{Remove low quality tasks}. We designed few-shot samples for agent to let it know what the low quality tasks look like, and prevent it from creating low quality tasks such as doing a very simple change like changing `counting apples' to `counting footballs'.
    \item \textbf{Remove easy tasks}. We have a difficulty review mechanism that can filter easy tasks out and keep difficult tasks. While building the benchmark, we conduct the review using five frontier T2I models: GPT-Image-2, Nano Banana 2, GPT-Image-1.5, FLUX.2 pro, and Seedream-4.5. A task is kept if at least two models receive a score, fewer than three models receive a full score of 100, and either the average score is below 35, or the average score is below 60 with at least one model receiving a score of 0. This keeps tasks that some models can solve but others fail, so our benchmark can better distinguish different models. 
\end{itemize}

Although we regenerate and evaluate these models after building the benchmark, difficulty filtering may still systematically lower the scores of the five models. To reduce this bias, we use five different frontier models from different model families, instead of using only one model or several models from the same family.

\begin{figure}[t]
    \centering
    \includegraphics[width=\textwidth]{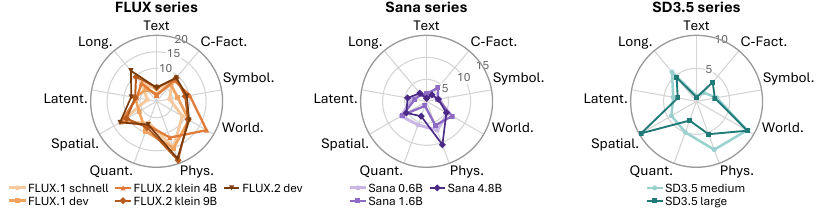}
    \caption{
        Performance of the FLUX series, Sana series and Stable Diffusion 3.5 series across nine axes. For these less strong open-source models, a newer or larger models' performance are not always better. Especially in some axes \emph{e.g.}, World knowledge grounding of FLUX series, Text Fidelity of Sana series, Physical consistency of Stable Diffusion 3.5, newer or larger models' performance getting worse instead.
    }
    \label{fig:open_series}
\end{figure}
\begin{table}[t]
\centering
\small
\caption{
Generation settings of open-source T2I models. Steps and guidance scale are the default values. The context length is the maximum number of tokens of the primary text encoder; we set it to the largest value the pipeline accepts, so that the full prompt is visible. All images are 1024$\times$1024.
}
\label{tab:open_source_settings}
\begin{tabular}{lcccc}
\toprule
\textbf{Model} & \textbf{Steps} & \textbf{Guidance} & \textbf{Context length} & \textbf{Precision} \\
\midrule
FLUX.2 dev      & 50 & 4.0 & 512  & bf16 \\
FLUX.2 klein 9B & 4  & 1.0 & 1024 & bf16 \\
FLUX.2 klein 4B & 4  & 1.0 & 1024 & bf16 \\
FLUX.1 dev      & 28 & 3.5 & 512  & bf16 \\
FLUX.1 schnell  & 4  & 0.0 & 512  & bf16 \\
Qwen-Image 20B  & 50 & 4.0 & 1024 & bf16 \\
SD3.5 large     & 28 & 3.5 & 512  & bf16 \\
SD3.5 medium    & 40 & 4.5 & 512  & bf16 \\
Sana 4.8B       & 20 & 4.5 & 512  & bf16 \\
Sana 1.6B       & 20 & 4.5 & 512  & fp16 \\
Sana 0.6B       & 20 & 4.5 & 512  & fp16 \\
\bottomrule
\end{tabular}
\end{table}

\begin{figure}[h]
    \centering
    \includegraphics[width=\textwidth]{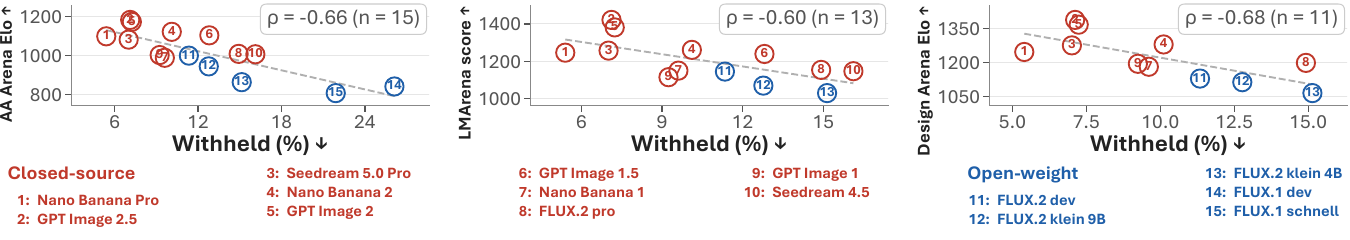}
    \caption{
    Correlation between withheld rate and human-preference arena scores. Higher arena scores correspond to lower withheld rates on GenOlympiad.
    }
    \label{fig:witheld_corr}
\end{figure}
\begin{figure}[h]
    \centering
    \includegraphics[width=\textwidth]{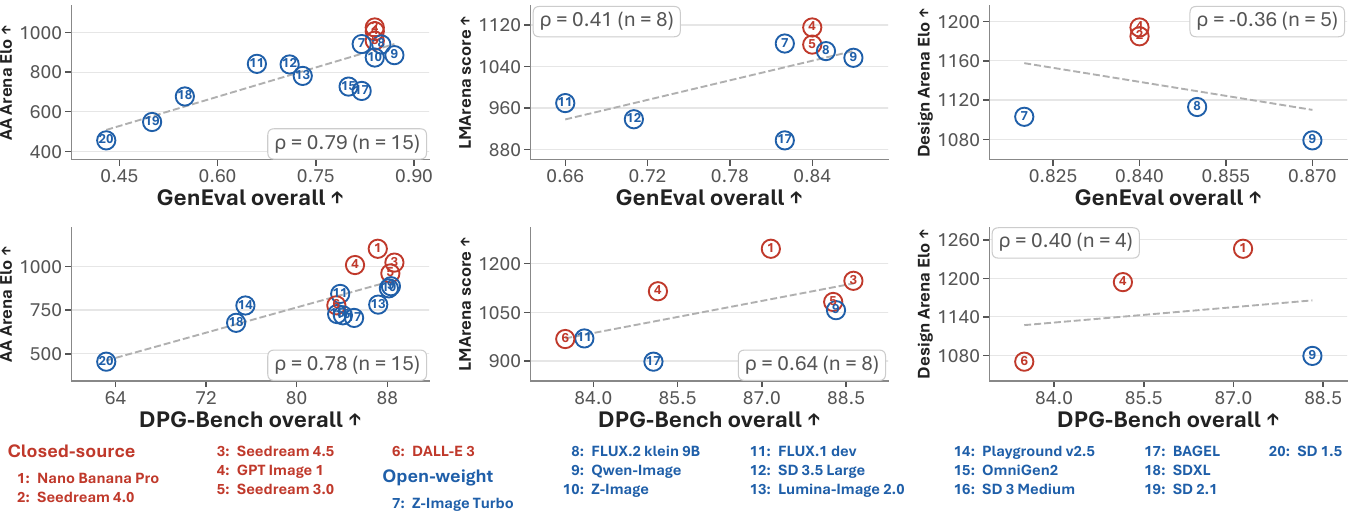}
    \caption{
    Correlation between arena and some other T2I benchmark GenEval and DPG-Bench. 
    }
    \label{fig:other_corr}
\end{figure}
\begin{figure}[h]
    \centering
    \includegraphics[width=\textwidth]{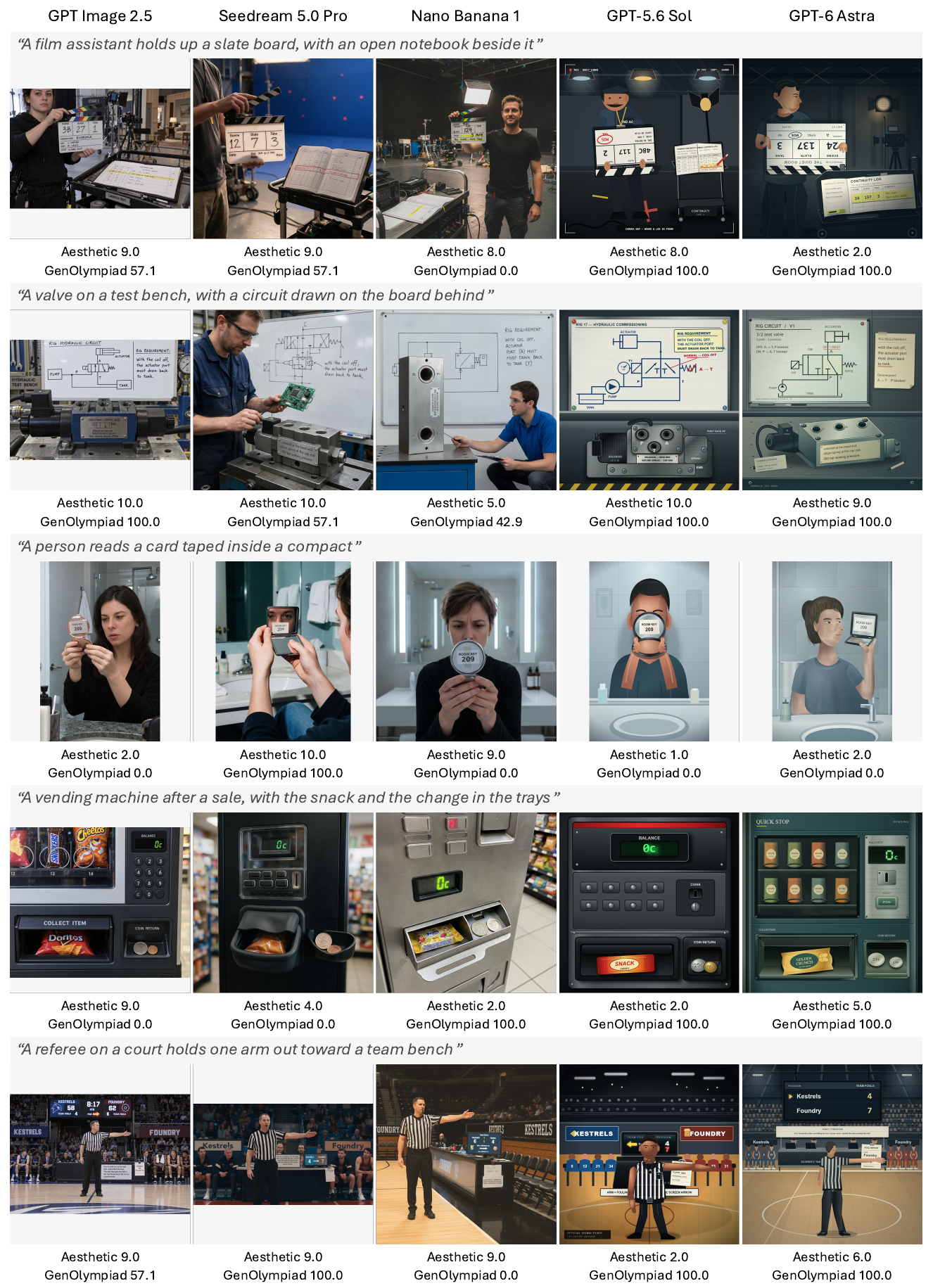}
    \caption{
    Sample images of each models with their aesthetic and GenOlympiad relaxed score. Prompts were shorten, not original.
    }
    \label{fig:aesthetic}
\end{figure}

\clearpage
\subsection{Prompt Templates}

\label{sec:prompt}
\raggedbottom
This section gives the prompts used in this paper, all prompts in abridged. We first give the few-shot examples that we wrote by hand before building the benchmark. We then give the instructions of the agents in the task generation protocol of Appendix~\ref{sec:howwebuild}. We evaluate coding models by giving system prompt. Some of them were written in Chinese and are translated here. All these following prompts are summarized and translated by Claude Fable 5.1.

\paragraph{Positive examples.} Each prompt below looks ordinary, but frontier T2I models fail it for a reason that can be named. Where we recorded the outcome of our first trials, it is given after the prompt.
\begin{enumerate}[leftmargin=*,itemsep=2pt]
\item \textbf{Exact geometry of a technical drawing.}
\begin{quote}\small
Generate a precise 2D CAD-style engineering drawing in strict orthographic front view of a flat mechanical panel. Visual style: - pure white background - thin black uniform lines - no perspective - no shading - no textures - no shadows - no color - no text - no dimension annotations - geometry only - crisp technical drawing appearance Geometry: 1. Outer boundary: a rectangle of width 300 units and height 200 units. 2. At the geometric center of the rectangle, draw two concentric circles: - outer circle diameter 60 units - inner circle diameter 30 units 3. Draw four identical circular holes of diameter 16 units, centered at: - (60, 150) - (240, 150) - (60, 50) - (240, 50) 4. Draw a horizontal slot centered at (150, 50), with total length 120 units and height 12 units, semicircular ends. 5. Draw two vertical rectangular slots, each width 12 units and height 70 units: - left slot center at (110, 100) - right slot center at (190, 100) 6. Draw a square cutout of side length 24 units centered at (150, 150). 7. The square cutout must be exactly centered horizontally with the concentric circles. 8. The two vertical slots must be symmetric about the vertical centerline of the plate. 9. All geometry must preserve exact proportions, alignment, symmetry, spacing, and relative size. Important: - The output must be a clean technical diagram. - All relative measurements must be preserved exactly. - Do not artistically reinterpret the drawing.
\end{quote}

\item \textbf{Two tilted mirrors.}
\begin{quote}\small
Two identical mirrors on a vanity, one tilted 10 degrees backward, one 10 degrees forward. A person at equal distance appears full-body in the backward-tilted mirror but cut off at the feet in the forward-tilted one.
\end{quote}

\item \textbf{Tree rings with an exact count.}
\begin{quote}\small
A polished 60cm tree-trunk cross-section with exactly 47 concentric annual rings, a label reading `Tree age: 47 years', a centimeter scale along the edge, plus two visibly narrow drought-ring clusters and one wide optimal-growth band at specific radii.
\end{quote}

\item \textbf{Falling Slinky.}
\begin{quote}\small
A 0.6m metallic Slinky released from rest is photographed mid-fall: the top coils have collapsed and fallen $\sim$3m while the bottom end remains motionless at the release height, the coils forming a compressed cluster whose spacing increases non-linearly from bottom to top.
\end{quote}

\item \textbf{Text printed on one grain of rice.}
\begin{quote}\small
A realistic macro photograph of a white ceramic plate filled with uncooked white rice grains. The camera is very close to the plate, with shallow depth of field, but the rice grains must remain individually visible. In the exact center of the image, there is one single rice grain lying horizontally on top of the pile. On this one rice grain, write the tiny black printed text: `GPT-IMAGE-2' The text must be physically printed directly on the surface of that rice grain, following the grain's curved shape. The letters must be extremely small but still clearly readable when zoomed in. Do not write any text on the plate, table, background, or other rice grains. Only the one central rice grain contains text. The image should look like a real macro lens photo, with natural lighting, realistic shadows, and a believable scale difference between the rice grain and the tiny text.
\end{quote}

\item \textbf{Three mutually consistent annotation panels.}
\begin{quote}\small
Generate a three-panel computer vision annotation benchmark image. The output must contain three equal-sized panels arranged horizontally from left to right, separated by narrow white gutters. Do not add titles, legends, or decorative elements. All three panels must represent the exact same scene with exactly eight apples on a kitchen table. All three panels must use the same camera view, crop, coordinate system, scale, and object layout. Panel 1: RGB image. A realistic RGB photograph of a kitchen table with exactly eight apples. The apples should vary in size, color, rotation, and pose. Some apples should touch or slightly overlap each other, but every apple must remain at least partially visible as a separate instance. Use realistic lighting, shadows, and perspective. Panel 2: instance segmentation mask. A clean pixel-aligned instance segmentation mask of Panel 1. Each of the eight apples must be represented by a different flat solid color, chosen from this fixed set: red, blue, green, yellow, magenta, cyan, orange, and purple. The table and background must be pure white. The mask must exactly match the visible shapes, boundaries, positions, scales, and occlusion relationships of the apples in Panel 1. Hidden parts of occluded apples must not appear in the mask. Do not include text, labels, outlines, shadows, gradients, or textures. Panel 3: object detection bounding box annotation. A clean annotation view of the same scene from Panel 1. Use a pure white background and draw exactly eight tight axis-aligned rectangular bounding boxes, one around each apple. Each bounding box must tightly enclose the full visible apple, including the visible stem if present. Each box must use the same color as the corresponding apple instance in Panel 2. Add a small numeric label from 1 to 8 at the top-left corner of each bounding box. Do not draw any extra boxes. Do not miss any apple. Do not render the apples themselves in Panel 3; show only the white background, the eight colored bounding boxes, and the numeric labels. Define instance identity consistently across all panels by sorting apples from left to right according to the x-coordinate of their visible-region centroids; if two apples have similar x-coordinates, break ties from top to bottom. The apple assigned identity 1 corresponds to label 1, identity 2 corresponds to label 2, and so on. The three panels must be mutually consistent. The same apple instance must have the same identity, position, size, overlap relationship, and visible extent across Panel 1, Panel 2, and Panel 3.
\end{quote}

\item \textbf{A chat screenshot whose picture contradicts its text.}
\begin{quote}\small
Generate a fictional AI chat screenshot used as a visual benchmark example. The chat interface shows a user prompt that confidently asks: `Draw a human hand with exactly five fingers. The final image must contain only five fingers.' The assistant response appears to comply and says: `Here is a realistic hand with exactly five fingers.' However, the generated image shown under the assistant response clearly depicts a realistic human hand with six fingers. The sixth finger should be visible but not exaggerated, making the failure subtle at first glance but undeniable upon inspection. Use a clean modern chat UI layout, rounded bubbles, white background, and a realistic image preview. The scene should demonstrate an instruction-following failure: the prompt asks for five fingers, but the output contains six. Do not use real ChatGPT logos, OpenAI logos, or any real platform branding. This must be a fictional mockup.
\end{quote}

\item \textbf{Offside at the moment of the pass.} \emph{(translated)}
\begin{quote}\small
Generate a realistic football match scene that captures the exact moment of a pass. The image must clearly show the passer, the receiver, at least three defenders, the direction of the goal, and the pitch markings. At the moment of the pass the receiver must be in an offside position: the receiver must be behind or level with the second-to-last defender. The ball must be clearly leaving the passer's foot at the moment of the pass.
\end{quote}

\item \textbf{Gomoku, one move from winning.} \emph{(translated)}
\begin{quote}\small
Generate a realistic top-down view of a gomoku game that shows a plausible position, in which every stone is placed with a tactical purpose. Do not place stones at random. Black needs only one move to win: one more black move makes five in a row, but in the current image Black must not already have five in a row. White should have several sensible defensive stones nearby that block or counter Black's attack. The stones elsewhere on the board should also be laid out sensibly, as in a real mid-to-late game. Render the board as a clean wooden gomoku board, with black and white stones placed exactly on the intersections, viewed from directly above.
\end{quote}

\item \textbf{A bar chart that describes its own image.} \emph{(translated)}
\begin{quote}\small
Generate a realistic office scene. On the wall there is a large whiteboard titled `Color distribution of this image'. The whiteboard shows a bar chart with four bars, labelled green, white, brown and blue. The heights of the four bars must accurately reflect the proportions of the pixels that are actually visible as green, white, brown and blue in the whole image. The room should contain green plants, white walls, brown wooden furniture and blue objects, so that all four colors are clearly present. The bar chart must be large, clearly legible and easy to measure visually.
\end{quote}

\item \textbf{Sheet music from LilyPond code.} \emph{(translated)}
\begin{quote}\small
Generate a clear black-and-white engraved sheet-music image from the following LilyPond code. The output must be a standard score excerpt on a white page, presented as a single treble staff. The clef, key signature, time signature, pickup measure, bar lines, pitches, durations, rests, ties and slurs must match the code exactly. Do not add extra notes, lyrics, fingerings, decorative graphics or unrelated annotations. The engraving should look like printed sheet music, not handwritten music.\par{\raggedright\ttfamily\footnotesize LilyPond code: \textbackslash{}version "2.24.0" \textbackslash{}paper \{ indent = 0\textbackslash{}mm line-width = 160\textbackslash{}mm tagline = \#\#f \} \textbackslash{}layout \{ ragged-right = \#\#t \} melody = \textbackslash{}relative c'' \{ \textbackslash{}clef treble \textbackslash{}key a \textbackslash{}minor \textbackslash{}time 3/8 \textbackslash{}tempo "Poco moto" \textbackslash{}partial 8 e16 dis | e16 dis e b d c | a8 r16 c e a | b8 r16 e gis b | c8 r16 e dis e | \} \textbackslash{}score \{ \textbackslash{}new Staff \textbackslash{}melody \}\par}
\end{quote}
The ground-truth image is rendered from the same code with LilyPond.

\item \textbf{Conway's Game of Life, first grid.}
\begin{quote}\small
Generate a clean black-and-white grid image showing the final state of Conway's Game of Life after 4 steps.\\
Use standard Conway's Game of Life rules on a finite grid with dead cells outside the boundary.\\
Black squares are live cells and white squares are dead cells.\\
Initial 10$\times$10 grid:\\
\texttt{0000000000}\\
\texttt{0000000000}\\
\texttt{0001000000}\\
\texttt{0000100000}\\
\texttt{0011100000}\\
\texttt{0000000000}\\
\texttt{0000000000}\\
\texttt{0000000000}\\
\texttt{0000000000}\\
\texttt{0000000000}\\
Render only the final 10$\times$10 grid after exactly 4 steps.\\
Draw each cell as a sharp square.\\
Do not add coordinates, labels, text, arrows, or decorative elements.
\end{quote}
The ground-truth image is computed with a Python program.

\item \textbf{Conway's Game of Life, second grid.}
\begin{quote}\small
Generate a clean black-and-white grid image showing the final state of Conway's Game of Life after 4 steps.\\
Use standard Conway's Game of Life rules on a finite grid with dead cells outside the boundary.\\
Black squares are live cells and white squares are dead cells.\\
Initial 12$\times$12 grid:\\
\texttt{000000000000}\\
\texttt{000001000000}\\
\texttt{000011100000}\\
\texttt{000000100000}\\
\texttt{000110000000}\\
\texttt{000101100000}\\
\texttt{000000000000}\\
\texttt{000001110000}\\
\texttt{000000010000}\\
\texttt{000000000000}\\
\texttt{000000000000}\\
\texttt{000000000000}\\
Render only the final 12$\times$12 grid after exactly 4 steps.\\
Draw each cell as a sharp square.\\
Do not add coordinates, labels, text, arrows, borders outside the grid, or decorative elements.
\end{quote}
The ground-truth image is computed with a Python program.

\item \textbf{Complex mirror and occlusion relations.} \emph{(translated)}
\begin{quote}\small
A view in front of a bathroom mirror. A bottle of shower gel stands on the counter. Behind the shower gel there is a bunch of keys. The keys are visible only in the mirror, because from the viewer's position they are hidden by the shower gel. The reflection in the mirror shows the complete back of the shower gel bottle. The reflection also contains an analog clock showing 11:07:47, and the picture in the mirror is mirrored.
\end{quote}
\emph{Observed: GPT-Image-2 fails; Seedream fails.}

\item \textbf{Most of the subject is not visible directly, but the whole is visible indirectly.} \emph{(translated)}
\begin{quote}\small
A street scene in realistic photo style. There is a small puddle on the ground in the foreground. On the right side of the frame only the front half of a bicycle can be seen directly: the front wheel, the handlebars and a short piece of the frame. The rear half and the rider's body are outside the edge of the frame or cropped, so the rider and the whole bicycle cannot be seen directly and completely in the main picture. However, in the reflection in the foreground puddle, the complete bicycle and the person riding it must be visible: both wheels, the whole frame, the saddle, and the rider's upper body and head. The complete bicycle and rider in the reflection must be consistent in position and orientation with the part of the bicycle that is visible in the main picture. The complete rider must not appear directly in the main picture, only in the puddle reflection. The whole image should look like a real street photograph.
\end{quote}
\emph{Observed: GPT-Image-2 fails; Seedream is correct.}

\item \textbf{A very large search space with an explicit global constraint.} \emph{(translated)}
\begin{quote}\small
Generate a picture of a maze. Only three paths lead from the start to the end. Mark these three paths in different colors.
\end{quote}
\emph{Observed: GPT-Image-2 fails; Seedream fails.}

\item \textbf{Complex topology.} \emph{(translated)}
\begin{quote}\small
A desk scene in realistic photo style. On the desk there are a black fountain pen and a pair of white wired earphones. The earphone cable is wound once around the pen. On the left side of the pen the cable passes over the pen and hides it; on the right side of the pen the cable goes under the pen and is hidden by it. The cable must be one continuous unbroken white line, and the pen must stay straight and must not be cut by the cable. The crossings must show a clear front-and-back relation and natural shadows.
\end{quote}
\emph{Observed: GPT-Image-2 fails; Seedream fails.}

\item \textbf{A large number that must be shown exactly.} \emph{(translated)}
\begin{quote}\small
A clean top-down photograph. Exactly 37 red M\&M chocolate candies are scattered on an ordinary wooden table. Each candy is clearly separated and can be counted individually. There is nothing else on the table.
\end{quote}
\emph{Observed: GPT-Image-2 fails; Seedream fails.}

\end{enumerate}
We also noted four failures without keeping the prompt text: dice whose pips should sum to a given number (the image summed to 58, which was wrong); a vernier caliper that should read 23.64 (wrong reading); a concave mirror magnifying a coin; and a plate of 20 grapes with a chart of them, where the chart was drawn wrong.

\paragraph{Examples added during construction.} While the benchmark was being built, we added five more examples that became the standard worked examples of the protocol. The idea of each example is ours. For three of them we show the final prompt of the task that was built from the idea. The other two were taught to the agent as ideas only: a spoken walking route that must be drawn as a floor plan, and the parts sheet of a layered 2D character that must assemble into the character.
\begin{enumerate}[leftmargin=*,itemsep=2pt]
\item \textbf{A reversed compass under the noon sun.}
\begin{quote}\small
Draw a clear picture of a hiker's brass novelty pocket compass lying on an open paper map on a wooden table. Show a bright SUN in the sky on the RIGHT side of the picture. It is SOLAR NOON in the northern hemisphere, so the sun marks the direction of geographic SOUTH. IMPORTANT INSTRUMENT RULE: this is a REVERSED novelty compass -- unlike a normal magnetic compass (whose red `N' needle points to geographic NORTH), on THIS compass the red `N' needle points to geographic SOUTH. Because the noon sun marks south, you must draw the red `N' needle pointing straight TOWARD the sun, and print the letter `N' on the rim of the rose on the side nearest the sun (with S opposite it, and E and W on the two sides). Make the sun a large clear icon and the north needle a distinct RED arrow tipped with the letter `N'. Do not add any other note or caption.
\end{quote}
The model tends to draw the compass that it has seen many times, with the red needle pointing away from the sun.

\item \textbf{The team at the bottom of the table celebrates.}
\begin{quote}\small
A photograph taken from the touchline of a small amateur football ground on a bright afternoon, after the last game of the day. Ordinary local club players, nobody famous, no sponsor logos.

A large painted RESULTS BOARD stands behind the pitch, headed `SATURDAY LEAGUE - RESULTS', and carries exactly these two result lines, one under the other, in big clear letters: SWIFTS 1 - 1 OTTERS OTTERS 2 - 0 BADGERS

Three clubs played today and each club's shirts are a plain single colour: the SWIFTS wear YELLOW, the OTTERS wear GREEN, the BADGERS wear PURPLE. Points are the usual three for a win and one for a draw.

This club has an odd tradition and the picture must show it: the CAPTAIN OF THE SIDE THAT COMES BOTTOM of that table stands on the grass with BOTH ARMS RAISED STRAIGHT ABOVE HIS HEAD, fists up, in an unmistakable victory pose; and the three SUBSTITUTES OF THE SIDE THAT COMES TOP sit on the bench slumped forward with their HEADS HELD IN BOTH HANDS, elbows on knees. Work the table out from the printed results yourself --- I am deliberately not telling you which side comes bottom or which comes top, and I am not telling you which shirt colour does what. Do not print the table, the points or any team's position anywhere in the picture: the only text in the picture is the board's heading and the result lines given above.

The results board, the celebrating captain and the three slumped substitutes on the bench must all be clearly visible in the same picture, and every shirt colour must be unmistakable.
\end{quote}
The rule is given in full, but which team comes bottom must be worked out from the printed results. The model tends to let the winning team celebrate.
\end{enumerate}

\paragraph{Negative examples.} Each prompt below looks hard, but GPT-Image-2 solves it, although other models may fail. For each one we wrote down why it is a poor task. The agent is told not to create tasks of these kinds.
\begin{enumerate}[leftmargin=*,itemsep=2pt]
\item \textbf{An agent can compute the result and then rewrite the prompt.} \emph{(translated)}
\begin{quote}\small
Generate an image: three empty cups A, B and C, each with a capacity of 100 ml. Tap A is off (0), tap B runs at 3 ml/s, tap C runs at 10 ml/s. After 8 seconds, draw the correct water level of each cup according to volume = flow rate $\times$ time (capped when the cup is full).
\end{quote}
\emph{Observed: GPT-Image-2 is correct; Seedream fails.}

\item \textbf{An agent can compute the result and then rewrite the prompt (second example).} \emph{(translated)}
\begin{quote}\small
A photorealistic illustration of a high-school physics problem. Take g = 10 m/s$^2$ and ignore air resistance. A small block A of mass 2 kg is released from rest at the top of a smooth incline of height 5 m. The bottom of the incline joins a rough horizontal track of length 4 m with a coefficient of kinetic friction $\mu$ = 0.25. After passing the rough track, the block makes a perfectly inelastic collision with a small block B of mass 3 kg that rests at the end of the track, and the two stick together. After the collision the two blocks move together onto a smooth horizontal track and finally fly off horizontally from the edge of a table of height 1.25 m. Draw the final state according to the laws of physics.
\end{quote}
\emph{Observed: GPT-Image-2 is correct; Seedream fails.}

\item \textbf{A language model alone can remove the difficulty.} \emph{(translated)}
\begin{quote}\small
A desk scene in realistic photo style. A white microwave oven stands in the middle of the desk. The camera faces the front of the microwave squarely, at the height of the centre of the microwave door. Only the front of the microwave, a little of the desk, and the plain wall behind can be seen; the back, top, left side and right side of the microwave cannot be seen. A very eye-catching cat-head-shaped sticky note is stuck in the middle of the back of the microwave. The note is bright pink, its edge has the outline of cat ears, the material is matte paper, the top-left corner curls up slightly, and it is held by a short piece of translucent tape. In the bottom-right corner of the note there are a small red heart and two yellow five-pointed stars, and on it, in clear black handwritten Chinese, is written: `Mom is not home today. Your favourite fried chicken is in the microwave.' The handwriting is neat, and the note is cute and eye-catching.
\end{quote}
\emph{Observed: GPT-Image-2 is correct; Seedream fails.}

\item \textbf{A language model alone can remove the difficulty (second example).} \emph{(translated)}
\begin{quote}\small
A desk scene in realistic photo style. A thick book lies in the middle of the desk, photographed from directly above and slightly in front. The front cover faces up, and the back cover faces down, flat against the desk. The front cover is pure white, with no text, pattern, label, decoration or stain. However, on the back cover there is a line of huge, striking red Chinese text: `Secret answer'. The text is printed in the middle of the back cover, in a very bold typeface and bright red, and is very eye-catching.
\end{quote}
\emph{Observed: GPT-Image-2 is correct; Seedream fails.}

\item \textbf{A simple change of topology, which GPT models handle easily.} \emph{(translated)}
\begin{quote}\small
A curved strip of paper bent into a half circle. On the outer side of the strip the word OUTSIDE is written repeatedly, and on the inner side the word INSIDE is written repeatedly. From the camera angle both the inner and the outer side can be seen, and each word must appear on the correct surface.
\end{quote}
\emph{Observed: GPT-Image-2 is correct; Seedream fails.}

\item \textbf{Complex relations among many objects, which GPT-Image-2 is good at.} \emph{(translated)}
\begin{quote}\small
A photorealistic still life that contains exactly eight objects, and all of the following relations must hold at the same time: a red cup is to the left of a blue book; a green ball is above a yellow box; an orange bottle is behind the blue book; a white bowl is in front of the red cup; a purple candle is to the right of the yellow box; a black phone leans against the orange bottle; a pink eraser is inside the white bowl; a brown wooden block is under the green ball. Apart from these eight objects, there must be nothing else.
\end{quote}
\emph{Observed: GPT-Image-2 is correct; Seedream fails.}

\item \textbf{The cause is not visible but its result is visible, which GPT models handle reliably.} \emph{(translated)}
\begin{quote}\small
A laptop stands on a desk, but from the current viewpoint the desk lamp behind the laptop cannot be seen. The lamp behind the laptop is the only light source. The laptop screen is off. There is a cat under the desk.
\end{quote}
\emph{Observed: GPT-Image-2 is correct; Seedream fails.}

\item \textbf{A precise gear ratio, which GPT models handle reliably.} \emph{(translated)}
\begin{quote}\small
Three meshing gears in a row: A (10 teeth), B (20 teeth) and C (40 teeth). Each gear has one red tooth pointing straight up. Gear A turns one full revolution. Show the final position of each red tooth. The positions of the three gears must fully obey the laws of physics. Label A, B and C (or left/right) clearly. The image must be photorealistic.
\end{quote}
\emph{Observed: GPT-Image-2 is correct; Seedream fails.}

\item \textbf{A counter-intuitive request that goes against physics, such as reversing the order of objects in a mirror.} \emph{(translated)}
\begin{quote}\small
A view facing a bathroom mirror. On the counter, from left to right: a red toothbrush, a blue cup, a bottle of yellow soap, and a white box labelled `Soap'. In the mirror the left-to-right order of these four objects is reversed, and the text on the box is mirrored too.
\end{quote}
\emph{Observed: GPT-Image-2 is correct; Seedream is correct.}

\end{enumerate}
From these trials we drew two lessons for the agent. First, a task is weak if its final picture can be described in words, because an agent can compute the result and rewrite the prompt into a simple description. Second, a task is weak if a language model can simplify the prompt, for example by dropping an object that the prompt asks for but that must not be visible.

\paragraph{Abridged instructions of the agents.} Each role in the protocol is played by a separate agent that works from a written instruction. The lists below are abridged translations of these instructions.

\providecommand{\agentbox}[2]{%
  \par\medskip\noindent
  \rule{\linewidth}{0.6pt}\par\nobreak\noindent
  {\small\textbf{#1} \emph{(abridged, translated from the original)}\par\nobreak
  \begin{enumerate}[leftmargin=2.4em,itemsep=0pt,topsep=2pt,parsep=1pt]
  #2
  \end{enumerate}}
  \par\nobreak\vspace{-2pt}\noindent\rule{\linewidth}{0.6pt}\par\medskip}

\agentbox{Instruction to the brainstorming agent}{
  \item Start from a real scene, such as a work station, a home, a street, a sport, a game or a document. Do not look for a scene to dress up a mechanism you already like.
  \item Inside the scene, find a hard constraint that a person who does this work would notice at a glance if it were drawn wrong. Only then ask whether it is difficult.
  \item Say in one sentence who would need this picture and what they would do with it. This field is required.
  \item Ask whether this picture existed in the world before we set the task. If not, the mechanism may still be good, but the carrier is a teaching prop: attach a real-life carrier and change nothing else.
  \item A task must satisfy four things at once: it is a real commission; a layman can tell right from wrong in one look; the answer appears nowhere in the prompt; and judging it leaves behind a picture worth showing.
  \item Tell your candidate in one sentence: what the viewer sees, where the mistake shows, and why this picture really exists. If you cannot, drop it.
  \item Take the difficulty from what actually breaks models: withhold the key number; make a quantity travel from one place of the picture to another; make each cell depend on other cells; ask for a composite transform to be inverted; leave one step outside the rule table; give no scale at all.
  \item Avoid what never breaks models: writing out the whole rule table for the model to follow; describing the final appearance; asking whether something appears anywhere in the image; exact counting on a regular layout.
  \item Difficulty must come from the process, not the dose. More objects, more rows, more decimal places and more arithmetic are a ladder, not a new task.
  \item To make a task harder, use a richer real specification or a real convention of the trade, or move the check to after the thing has been assembled. The amount of data stays the same.
  \item One conversion, one addition or one comparison is allowed on any axis. Multi-step arithmetic belongs to the latent computation axis and may not be disguised.
  \item Test the candidate three times. Hand the withheld number to the model for free: if the task becomes easy, it tests arithmetic. Reduce the count to five: if it becomes easy, it tests counting. Make the unusual rule normal again: if it is still hard, the difficulty is not where you said it was.
  \item One task per mechanism. Two candidates are the same task if their failure conditions are the same sentence, whatever the objects are. Check for collisions across all nine capability axes, not only your own.
  \item Do not clone the worked examples. Learn the formula, not the picture.
  \item Before a candidate is released, name the nearest task that has already been scored, give its score, and say which specific quantity this candidate keeps outside the prompt. `It is harder' is not an answer.
  \item Declare anything you truncated, sampled or relaxed. You do not approve your own candidate: a reviewer does.
}

\agentbox{Instruction to the task-writing agent}{
  \item Write the failure condition and the evaluation program before you write the prompt.
  \item The prompt is a work order, not a construction drawing. It has four parts: the scene and the deliverable; the material that fixes the inputs; the purpose; and the constraints, including what must not appear. Never teach the model how to draw, and never state the answer.
  \item Fix the input, not the output. Whatever the model should have before it starts may be given exactly. Whatever it should only produce after working may not appear.
  \item List every element of the expected answer and confirm that at least one cannot be found word for word in the prompt.
  \item A full rule table may be given, but after the table there must still be one step of computation, placement or conservation that cannot be looked up in it.
  \item Do not lay the picture out element by element with a string of imperatives. That turns the model into a typesetter and measures nothing. Do not write down the final appearance.
  \item The expected answer must promise something, and it must contain numbers.
  \item A real industrial setting is not enough by itself. Its value is the part of the real specification that cannot fit in the prompt. Say which part that is.
  \item For a task about orientation or visibility, fix the orientation with a natural anchor in the scene, such as a person reading the face of the object. Never state the camera-side conclusion, and never stay silent about orientation.
  \item Rotate your own reference image by thirty degrees and score it again. If the score flips, the orientation was not fixed hard enough. State the sign convention wherever one shape could be read as two different signs.
  \item If the VLM could answer the question from its own knowledge without looking at the image, the candidate is rejected. No rewording rescues it.
  \item Answer two questions in writing. How could a model that knows nothing guess a score? How could a strong model get around the ability itself? Close both routes.
  \item Require a result that cannot be faked by drawing, such as specific numbers or a correctly placed object. Do not ask whether something appears anywhere in the image.
  \item No exact counting on a structured layout such as a table or a calendar. Counting tasks must scatter their objects.
  \item Across the difficulty levels of one task, change at least two of carrier, setting, composition and style. Changing only the numbers is not enough. Re-skinning counts only if it changes the mechanism: counting apples and counting pears is one task, not two.
  \item The picture that the prompt describes must have existed in the world before the task was set.
}

\agentbox{Instruction to the evaluation-program agent}{
  \item Perception belongs to the VLM and arithmetic belongs to code. The VLM never does arithmetic. Where the answer needs a computation, the VLM only reports what it sees in a machine-readable form, and a program computes the conclusion.
  \item Every step must leave an overlay on the image that marks the element this step looked at.
  \item Ask the VLM one observable fact at a time. Every `and' belongs in code. Split a compound question into separate questions and combine them in code.
  \item Every question must be answerable in seconds by a person with no background who is shown the crop. Move all domain knowledge into code. Only two kinds of question may need expertise: transcribing a notation, and marking geometry.
  \item One question, one fact. No comparisons that are hard to follow, no double negatives, and no examples, because an example becomes the VLM's target list. Fix the answer format.
  \item Keep the phrase that names what to locate short. A long phrase breaks the check itself, and the symptom is a refusal to read, not a wrong number.
  \item A question may never contain a correctness word, state an expectation, or reveal the shape of the expected answer. Never give the expected count to the step that proposes the boxes. The count is the answer.
  \item Do not add a scene-level easy question by default. Only a real precondition earns one, because a scene check tends to reject the most correctly drawn images of an unusual task.
  \item If the expected answer fixes the geometry, draw it over the image and ask whether it coincides. If the criterion is a geometric relation, draw the construction so that the conclusion is visible on the image.
  \item A count that decides the score must go through the loop: locate, confirm, read, compute. The scored count is the number of items confirmed one by one, never the number of boxes proposed. Counts of at most ten are exempt.
  \item Use only the allowed operators. A new operator needs a task-neutral design and at least three tasks that will reuse it.
  \item Complexity is a defect. A task that no allowed operator can evaluate is better removed than kept.
  \item There are three outcomes, never two: a scored result; a zero caused by the model's own broken image; and a failure of the evaluator. A failure of the evaluator, or a refusal by the VLM, withholds the task and is never scored as zero.
  \item Every zero must name the fact that was violated. That fact must be drawn on the image, and it must have been stated in plain words in the prompt.
  \item Never change a threshold because of a score you have seen. The only valid reasons are the measured noise of the evaluator, consistency with similar tasks, and the arithmetic of the task itself.
}

\paragraph{System prompt of the coding models.} GPT-6 Astra and GPT-5.6 Sol receive the system prompt below and the task prompt as the user message. \{W\} and \{H\} are replaced by the width and the height of the image in pixels.
\begin{quote}\small
You draw pictures by writing code.

The user message is a picture description. Write one self-contained HTML document that draws that picture, and return nothing else.

\textbf{Looking Things Up.}
\begin{itemize}[leftmargin=*,itemsep=0pt,topsep=1pt]
\item You have a web search tool. Use it whenever the description depends on a fact you would otherwise be guessing at: a live or dated figure, the real appearance, markings, proportions or specification of an actual object, place, instrument or document. Search first, then draw what you found.
\item Search is yours to judge. A picture that needs no outside fact needs no search.
\item What you bring back from a search must end up in the code you write. The rendered page itself has no network: it cannot load a photograph, a font, a map tile or a data file from anywhere, so a fact is only useful once you have turned it into drawing instructions. Pasting a picture from the web would be collecting, not drawing, and the page cannot do it anyway.
\end{itemize}

\textbf{Canvas.}
\begin{itemize}[leftmargin=*,itemsep=0pt,topsep=1pt]
\item The drawing surface is exactly \{W\} pixels wide and \{H\} pixels high. Fill it. Do not leave a border of blank page around the picture and do not draw outside it.
\item Use either a \textless{}canvas\textgreater{} element of exactly \{W\}x\{H\} with an inline \textless{}script\textgreater{} that draws on its 2D context, or one inline \textless{}svg\textgreater{} element of exactly \{W\}x\{H\}. Either is fine. Mixing them is fine.
\item Give the page a background. A transparent or default-white page counts as an undrawn picture.
\end{itemize}

\textbf{Self-Contained, No Network.}
\begin{itemize}[leftmargin=*,itemsep=0pt,topsep=1pt]
\item Everything is inline in the one document. No \textless{}img src\textgreater{}, no \textless{}link\textgreater{}, no external \textless{}script src\textgreater{}, no fetch, no XMLHttpRequest, no WebSocket, no import from a URL, no @font-face pointing at a file, no web font service. Nothing is downloaded: the page is rendered with the network off, so anything external simply will not arrive.
\item Text must use fonts already on a normal machine. Name generic families such as sans-serif, serif, monospace, or common system faces with a generic fallback.
\item No alert, confirm or prompt. No window.open. No timers that keep changing the picture: draw the finished picture, do not animate it.
\end{itemize}

\textbf{When The Picture Is Finished.}
\begin{itemize}[leftmargin=*,itemsep=0pt,topsep=1pt]
\item Draw on load, synchronously if you can.
\item The moment the drawing is complete, set window.\_\_done = true. A screenshot is taken as soon as that flag appears, so setting it early gives you a half-drawn picture and never setting it costs you nothing but a wait.
\item The document must not throw. An uncaught exception is a blank page and scores nothing.
\end{itemize}

\textbf{What You Are Being Asked For.}
\begin{itemize}[leftmargin=*,itemsep=0pt,topsep=1pt]
\item Draw everything the description asks for, with the counts, the positions, the text, the relations and the values it names. Every quantity in the description is checkable in the picture afterwards, so compute it in code rather than eyeballing it: lay out with arithmetic, place things by coordinates you work out, and write the labels the description dictates, spelled exactly as it spells them.
\item You get one attempt and you never see the rendered result, so prefer constructions you can be sure of over constructions that need tuning by eye.
\item Make it look like the thing described, not like a diagram of the thing, unless a diagram is what was described.
\end{itemize}

\textbf{Output Format.}
\begin{itemize}[leftmargin=*,itemsep=0pt,topsep=1pt]
\item Reply with exactly one fenced code block, opened with three backticks and the word html, and nothing before or after it. No explanation, no notes, no second block.
\end{itemize}

\end{quote}

\end{document}